\documentclass[10pt,twocolumn,letterpaper]{article}

\usepackage[pagenumbers]{cvpr} 

\usepackage[T1]{fontenc}

\usepackage{cuted}    
\usepackage{currfile} 
\usepackage{caption}  

\usepackage{graphicx}
\usepackage{amsmath}
\usepackage{amssymb}
\usepackage{booktabs}
\usepackage{colortbl} 
\definecolor{rowhighlight}{RGB}{234,245,242}
\usepackage{tabularx}
\usepackage{multirow}
\usepackage{xspace}

\ExplSyntaxOn
\cs_new_protected:Npn \__paper_sout:n #1 {
  \leavevmode
  \seq_set_split:Nnn \l_tmpa_seq { ~ } { #1 }
  \seq_map_indexed_inline:Nn \l_tmpa_seq {
    \int_compare:nNnT { ##1 } > { 1 } { \space }
    \hbox_set:Nn \l_tmpa_box { ##2 }
    \hbox_overlap_right:n { \rule[0.5ex]{\box_wd:N \l_tmpa_box}{0.4pt} }
    \box_use:N \l_tmpa_box
  }
}

\ExplSyntaxOff
\IfFileExists{ulem.sty}{\usepackage[normalem]{ulem}}{}

\newcommand{\R}{\mathbb{R}}

\newcommand{\method}{Spectator\xspace}           
\newcommand{\bench}{BrainStateBench\xspace}      

\makeatletter
\newcommand{\compactmainspacing}{%
  \raggedbottom
  \setlength{\textfloatsep}{14pt plus 2pt minus 2pt}%
  \setlength{\dbltextfloatsep}{14pt plus 2pt minus 2pt}%
  \setlength{\floatsep}{10pt plus 2pt minus 2pt}%
  \setlength{\dblfloatsep}{10pt plus 2pt minus 2pt}%
  \captionsetup{skip=7pt}%
  \renewcommand{\paragraph}{\@startsection{paragraph}{4}{\z@}%
    {14pt plus 1pt minus 1pt}{-1em}{\normalfont\normalsize\bfseries}}%
}
\makeatother

\usepackage{tikz}
\newcommand{\globeicon}{\raisebox{-0.15em}{\tikz[line width=0.07em, x=0.95em, y=0.95em, cvprblue]{
  \draw (0.5,0.5) circle (0.5); \draw (0.5,0.5) ellipse (0.22 and 0.5); \draw (0,0.5) -- (1,0.5);
  \draw (0.067,0.75) -- (0.933,0.75); \draw (0.067,0.25) -- (0.933,0.25);}}}

\providecommand{\leftlinenumbers}{}
\providecommand{\linenumbers}{}
\providecommand{\nolinenumbers}{}
\ifdefined\linenumbersep\else\newlength{\linenumbersep}\fi

\definecolor{cvprblue}{rgb}{0.21,0.49,0.74}
\usepackage[pagebackref,breaklinks,colorlinks,allcolors=cvprblue]{hyperref}

\def\confName{CVPR}
\def\confYear{2027}

\title{State of Mind: Learning and Evaluating World Models of Naturalistic Brain Dynamics}

\author{
Yuxiang Wei$^{1,2,*}$ \qquad Zundong Wu$^{1,2,*}$ \qquad Weiyu Sun$^{1,*}$ \\
Qizhen Lan$^{3}$ \qquad Yanteng Zhang$^{2}$ \qquad Vince D. Calhoun$^{2}$ \\[0.5em]
{\normalsize $^1$Georgia Institute of Technology \qquad $^2$Center for Translational Research in Neuroimaging and Data Science} \\
{\normalsize $^3$University of Texas Health Science Center at Houston} \\[0.5em]
{\normalsize \globeicon\ \href{https://yuxiangwei0808.github.io/State-of-Mind/}{yuxiangwei0808.github.io/State-of-Mind}} \\[0.3em]
{\small $^*$Equal contribution}
}

\begin{document}
\begingroup
\compactmainspacing
\maketitle
\begin{strip}
\centering
\includegraphics[width=\linewidth]{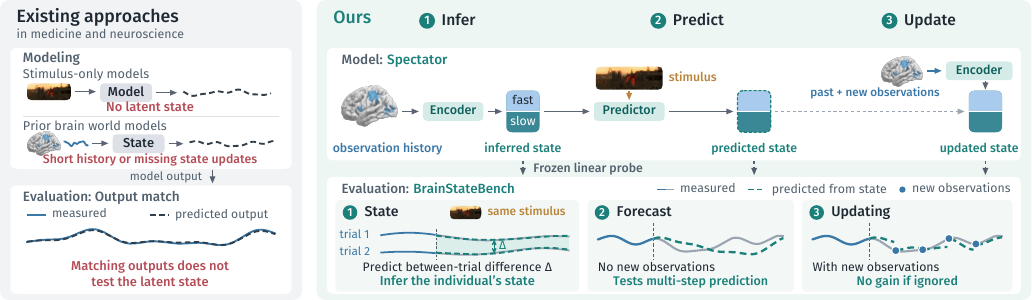}
\captionsetup{hypcap=false}
\captionof{figure}{\textbf{Modeling and evaluating individual brain states.} \emph{Left:} stimulus-only models maintain no latent state, while prior brain world models use short histories or lack state updating (\cref{sec:related}). Judging models solely by prediction accuracy conflates stimulus-driven responses with state modeling. \emph{Right:} \method infers, predicts, and updates a two-timescale latent state to track an individual's brain state. Moving beyond standard metrics, \bench probes a model's latent state directly via \emph{State}, \emph{Forecast}, and \emph{Updating}.}
\captionsetup{hypcap=true}
\label{fig:teaser}
\end{strip}

\begin{abstract}
Understanding naturalistic brain activity requires going beyond the stimulus alone, as external inputs continuously interact with an individual's evolving brain state.
A world model captures such dynamics in a latent state by inferring it from neural observations, predicting stimulus-driven evolution, and updating with new observations.
However, existing brain world models fail to maintain a persistent latent state over time, and standard metrics of neural prediction accuracy cannot tell whether a model tracks an individual's brain state or relies solely on the stimulus.
To directly test state inference, prediction, and updating, we introduce \textbf{\bench}, which evaluates a model's latent state through a frozen linear probe on whole-cortex, movie-driven fMRI.
Its \emph{State} task tests whether a model captures trial-to-trial variability in an individual's responses to repeated stimuli, while \emph{Forecast} and \emph{Updating} evaluate future activity predictions without and with new observations, respectively. 
To track individual brain state across time, we propose \textbf{\method}, a linear latent world model trained with a joint-embedding predictive objective.
\method encodes short- and long-horizon observation histories into a two-timescale latent state and predicts state transitions linearly from the current state and stimulus.
Stimulus-only encoders reach 0.320 in fMRI prediction correlation without observing the brain (0\% on \emph{State}), while prior deep brain world models reach 0.239 yet score at most 0.4\% on \emph{State}.
In contrast, \method achieves 0.349 correlation alongside 5.5\% on \emph{State}, 16.0\% on \emph{Forecast}, and 20.3\% on \emph{Updating}, outperforming all baselines while forecasting over $20\times$ faster than deep forecasting models.
Decomposing \method's forecasts shows that inferred state drives early predictions across cortex, peaking in association areas, while the stimulus dominates later predictions in shared sensory and narrative-processing regions, establishing brain world models as a framework for disentangling how internal state and external stimuli shape individual brain activity.
\end{abstract}
\section{Introduction}
\label{sec:intro}

Brain activity during a naturalistic stimulus arises from a continuous interaction between the incoming stimulus and the individual's evolving brain state~\cite{hasson2004intersubject,fox2006coherent}.
Because of these ongoing internal dynamics, even an identical stimulus evokes different responses across individuals and across repeated trials within the same individual, which the stimulus alone cannot explain: an individual watching the same movie twice is in a different state of mind each time. Capturing naturalistic brain dynamics therefore requires inferring this unobserved brain state from past neural activity and tracking how it evolves under incoming stimuli. Doing so makes it possible both to predict an individual's future responses and to study how internal dynamics and external inputs jointly shape brain activity.

World models provide a principled formulation for this goal: they infer a latent state from observations, predict how it evolves under actions, and update it as new observations arrive~\cite{kalman1960,ha2018worldmodels,lecun2022path,hafner2025dreamerv3}.
In neural modeling, the stimulus acts as the action, fMRI recordings provide the observations, and the latent state represents the individual's brain state.
Yet existing approaches fall short on both modeling and evaluation (\cref{fig:teaser}, left).
On the \textbf{modeling side}, stimulus-only encoding models predict shared stimulus-driven activity without observing the individual's brain and maintain no latent state~\cite{dascoli2026tribev2,gokce2026mirage}. Medical world models learn image representations, guide imaging, and simulate treatment outcomes~\cite{yue2025chexworld,yue2025echoworld,yang2025mewm}. Existing brain world models~\cite{dong2026neuroworld} and fMRI forecasters~\cite{yin2026brainvista,gao2026braincast} either forecast directly in observation space or encode only brief observation windows without maintaining a persistent, updatable latent state. On the \textbf{evaluation side}, standard metrics score fMRI prediction accuracy, mixing shared stimulus-driven responses with internal dynamics so one cannot tell whether a model tracks an individual's brain state or simply predicts the stimulus response.

To directly test state inference, prediction, and updating, we introduce \textbf{\bench} on whole-cortex, movie-driven fMRI (\cref{fig:teaser}, right).
Rather than scoring end-to-end fMRI predictions alone, \bench maps each model's latent state to fMRI via a linear probe fitted on training data and then frozen, evaluating diverse architectures on equal footing without retraining.
Its \emph{State} task pairs two trials of the same stimulus from the same individual and tests whether the probed state predicts the response difference between them, where any stimulus-only model scores zero by construction.
\emph{Forecast} evaluates one-minute predictions of future activity without new observations, while \emph{Updating} tests how effectively a model revises its predictions as new observations arrive.

To maintain and update a persistent latent state over time, we propose \textbf{\method}, a linear latent world model of naturalistic brain dynamics (\cref{fig:teaser}, right; \cref{fig:model}).
Its encoder compresses observation history into a two-timescale latent state capturing fast dynamics over seconds and slow context over minutes, while its predictor advances this state using incoming stimulus features.
When new observations arrive, they pass through the encoder to update the latent state. Following JEPA~\cite{lecun2022path,assran2023ijepa}, we train the encoder and predictor by aligning multi-step predicted latent states with target states inferred from future observations.
Crucially, because its dynamics are linear, every forecast decomposes exactly into additive latent-state and stimulus terms, making their contributions directly interpretable.

Experiments show why brain world models should be evaluated on their latent state (\cref{sec:main,sec:inputs}).
Stimulus-only encoders reach 0.320 in fMRI prediction correlation without neural input (0\% on \emph{State}), while prior brain world models reach 0.239 yet capture almost no trial-specific state (at most 0.4\% on \emph{State}). Across multi-minute horizons, \method outperforms all six baselines across every metric, achieving 0.349 correlation alongside 5.5\% on \emph{State}, 16.0\% on \emph{Forecast}, and 20.3\% on \emph{Updating}, more than double the best baseline. Ablations show that its learned slow state drives long-horizon gains, while linear dynamics beat Transformer variants and remain stable across five-minute rollouts (\cref{sec:ablations}).
Moreover, \method forecasts over $20\times$ faster than prior forecasting models, matches baselines trained on full data using 10--25\% of recordings, and generalizes across scanners to unseen individuals on a second dataset without calibration (\cref{sec:main}). 
Finally, decomposing \method's forecasts reveals a clear cortical hierarchy: inferred latent state drives early predictions, especially in prefrontal and parietal cortex, whereas the stimulus dominates later predictions in sensory and narrative-processing regions (\cref{sec:analysis}).

Our contributions are:
\begin{itemize}
\item \textbf{State-centric evaluation of brain world models.}
We introduce \textbf{\bench}, moving beyond prediction accuracy to directly probe latent state inference, prediction, and updating.

\item \textbf{A simple and effective linear brain world model.}
We propose \textbf{\method}, showing that linear latent dynamics over multi-minute horizons outperform complex nonlinear and stimulus-only models across tasks and datasets at over $20\times$ the speed.

\item \textbf{Insights into individualized brain dynamics.}
We show that \method naturally disentangles shared stimulus-driven responses from an individual's trial-specific internal brain state, revealing where and when each governs naturalistic brain activity.
\end{itemize}

\section{Related work}
\label{sec:related}

\paragraph{World models and their evaluation.}
World models learn dynamics for prediction and planning~\cite{ha2018worldmodels,hafner2025dreamerv3,bruce2024genie,yang2024unisim}.
Latent world models predict in representation space, with JEPA learning encoders and predictors by matching predicted and target representations~\cite{lecun2022path,assran2023ijepa,bardes2024vjepa}.
Linear state-space models provide a classical formulation of latent dynamics~\cite{kalman1960}, including input-driven neural activity~\cite{vahidi2024ipsid}.
\method combines linear dynamics with a JEPA objective to learn a two-timescale brain state.
Evaluation is a separate challenge: output quality does not test a model's grasp of the underlying dynamics~\cite{shang2026worldarena,zhang2026worldinworld}.
Existing work probes learned representations~\cite{li2023othello,assran2025vjepa2} or compares predicted trajectories with known system states~\cite{vafa2024evaluating}.
For the brain, \bench tests latent-state capabilities through neural measurements because the underlying state is unobserved.

\paragraph{World models of brain dynamics.}
Medical world models span diagnosis, imaging, and treatment simulation~\cite{yue2025chexworld,yue2025echoworld,yang2025mewm}.
For brain dynamics, NeuroWorld learns stimulus-conditioned latent transitions and decodes rollouts into fMRI~\cite{dong2026neuroworld}.
BrainVista predicts brain activity autoregressively from network-wise tokens and multimodal stimuli~\cite{yin2026brainvista}.
Other approaches generate whole-brain fMRI conditioned on anatomy~\cite{xia2026brainworld}, forecast from neural history~\cite{gao2026braincast,rogalla2026afm}, or learn representations for classification~\cite{dong2024brainjepa}.
Brain forecasting and generation models are evaluated by how well their outputs match measured activity.
\method integrates multi-minute neural history into a latent state that evolves under stimuli and incorporates new observations, while \bench evaluates state inference, prediction, and updating.
\Cref{sec:app-related} discusses broader medical applications and world-model evaluation.

\paragraph{Brain encoding models and benchmarks.}
Encoding models predict brain activity from stimulus features~\cite{naselaris2011encoding,huth2016semantic}.
Recent models trained on movie-fMRI datasets accurately predict stimulus-driven activity using video, audio, and language features~\cite{dascoli2025tribe,dascoli2026tribev2,gokce2026mirage}.
Because they receive no neural observations at test time, they do not infer the individual's ongoing brain state and serve as stimulus-only references in \bench.
Brain-Score~\cite{schrimpf2020brainscore} and Algonauts~\cite{gifford2025algonauts} assess alignment between model predictions and measured brain responses.
\bench asks what a latent state contributes beyond the stimulus, including differences between repeated trials and the benefit of new neural observations.

\section{\method: A Linear Latent Brain World Model}
\label{sec:model}

\method is a latent world model of naturalistic brain dynamics (\cref{fig:model}).
An encoder compresses recent neural observations into a two-timescale latent state, and a predictor evolves this state forward under incoming stimulus features.
Rather than forecasting noisy fMRI directly, \method rolls out dynamics entirely in latent space, aligning predicted trajectories with future states inferred by a target encoder.

\begin{figure}[t]
\centering
\includegraphics[width=\linewidth]{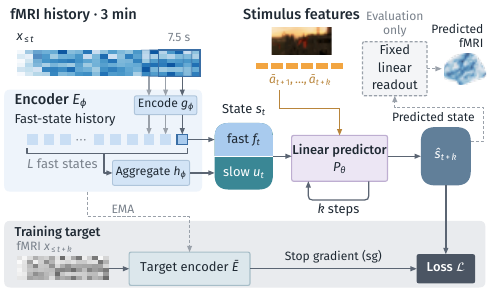}
\caption{\textbf{\method architecture and training.} The encoder maps recent fMRI observations to a fast state $f_t$ and aggregates fast-state history into a slow state $u_t$, forming the latent state $s_t = [f_t; u_t]$. A linear predictor evolves this state forward with multimodal movie features. During joint-embedding training, multi-step predicted states are aligned with stop-gradient target states inferred from future observations by an exponential moving average (EMA) target encoder $\bar E$. A frozen linear probe (readout) maps latent states to whole-cortex fMRI strictly for evaluation (\cref{sec:probe}).}
\label{fig:model}
\end{figure}

\subsection{Problem formulation}
\label{sec:setting}
Consider an individual experiencing a continuous naturalistic stimulus.
At each discrete time step $t$, we observe neural activity $x_t \in \R^D$ (whole-cortex fMRI parcellated into $D = 1{,}000$ regions~\cite{schaefer2018parcellation}) alongside stimulus features $a_t$ extracted from the movie's video, audio, and transcript by a pretrained multimodal encoder~\cite{dascoli2026tribev2}.
Treating stimulus features as external actions, a latent brain world model comprises an encoder $E_\phi$ and a predictor $P_\theta$:
\begin{align}
\text{Encoder:} \quad & s_t = E_\phi(x_{\le t}), \\
\text{Predictor:} \quad & \hat s_{t+1} = P_\theta(\hat s_t, a_{\le t+1}).
\end{align}
Together, $E_\phi$ and $P_\theta$ support the three core capabilities of a world model.
\emph{Inference} maps observations up to step $t$ into the current latent state $s_t$.
\emph{Prediction} initializes $\hat s_t = s_t$ and autoregressively rolls out $\hat s_{t+1}, \hat s_{t+2}, \dots$ driven solely by incoming stimulus features, requiring no neural observations after step $t$.
\emph{Updating} incorporates newly arrived observations through $E_\phi$ to refresh the latent state.

\subsection{Two-timescale encoder}
\label{sec:encoder}
Cortical responses to naturalistic stimuli unfold across a hierarchy of timescales: sensory areas track immediate inputs within seconds, whereas higher-order regions integrate context over tens of seconds to minutes~\cite{hasson2008hierarchy,yeshurun2021dmn}.
To capture both regimes, the encoder constructs a two-timescale latent state $s_t = [f_t; u_t]$ comprising a fast state $f_t$ and a slow state $u_t$.
A fast network $g_\phi$ extracts $f_t$ from the $m$ most recent observations,
\begin{equation}
f_t = g_\phi(x_{t-m+1}, \dots, x_t),
\end{equation}
while a slow network $h_\phi$ summarizes fast-state history over a longer horizon of $L$ steps:
\begin{equation}
u_t = h_\phi(f_{t-L+1}, \dots, f_t).
\end{equation}

\subsection{Residual latent predictor}
\label{sec:predictor}
The predictor models latent state transitions via a residual update $p_\theta$ conditioned on the current state and stimulus:
\begin{equation}
\hat s_{t+1} = \hat s_t + p_\theta(\hat s_t, \bar a_{t+1}),
\label{eq:predictor}
\end{equation}
where $\bar a_{t+1}$ concatenates stimulus features from recent steps up to $t+1$, allowing the predictor to account for the delay between stimulus onset and the recorded neural response (e.g., the hemodynamic lag in fMRI). During a rollout, $p_\theta$ advances the fast and slow states together, and $h_\phi$ is applied only inside the encoder.

\subsection{Two-stage training}
\label{sec:training}
Training proceeds in two stages: observation-only initialization followed by joint-embedding latent prediction.

\paragraph{Encoder pretraining.}
We first pretrain the encoder on neural observations alone to initialize informative state representations.
Specifically, $g_\phi$ is trained alongside an auxiliary predictor and observation readout---both discarded after pretraining---to forecast upcoming observations from the preceding $m$ steps.
We then fit $h_\phi$ so that the slow state $u_t$ predicts the mean of future fast states.

\paragraph{Latent prediction.}
Next, we jointly train $E_\phi$ and $P_\theta$ under a multi-step JEPA objective~\cite{lecun2022path,assran2023ijepa}, requiring open-loop state rollouts to match future inferred states.
Starting from $\hat s_t = s_t$, the predictor unrolls $K$ steps via \cref{eq:predictor} to produce $\hat s_{t+1}, \dots, \hat s_{t+K}$, minimizing the multi-step squared error:
\begin{equation}
\mathcal{L} = \sum_{t} \sum_{k=1}^{K} \bigl\| \hat s_{t+k} - \mathrm{sg}\bigl(\bar E(x_{\le t+k})\bigr) \bigr\|_2^2.
\label{eq:loss}
\end{equation}
Following~\cite{grill2020byol,assran2023ijepa}, the target encoder $\bar E$ mirrors $E_\phi$ with weights updated as an exponential moving average (EMA) of $\phi$, and $\mathrm{sg}(\cdot)$ denotes the stop-gradient operator.

\subsection{Linear instantiation}
\label{sec:linear}
While the formulation above admits nonlinear architectures such as Transformers (\cref{sec:experiments}), \method instantiates $g_\phi$, $h_\phi$, and $p_\theta$ as structured linear operators:
\begin{align}
f_t &= W_f\,[x_{t-m+1}; \dots; x_t], \\
u_{t,i} &= \sum_{\ell=0}^{L-1} w_{i\ell}\, f_{t-\ell,i}, \\
p_\theta(s, \bar a) &= W\,[s;\ \bar a;\ 1],
\end{align}
where each slow-state dimension $i$ aggregates its past fast states across lags $\ell$ via learned per-channel weights $w_{i\ell}$, and $[s;\ \bar a;\ 1]$ appends a unit bias.

This linear design yields two key advantages. First, the final predictor can be fitted in closed form: after latent-prediction training, we freeze the encoder and solve for $W$ over the training data via ridge regression,
\begin{equation}
W = \arg\min_{W} \sum_t \bigl\| s_{t+1} - s_t - W\,[s_t;\ \bar a_{t+1};\ 1] \bigr\|^2 + \lambda \| W \|^2.
\end{equation}
Second, whereas nonlinear predictors entangle state and stimulus across rollout steps, linear dynamics express each forecast as the exact sum of an initial-state term and a stimulus term.
Because the evaluation probe is also linear, we can directly isolate or swap state and stimulus contributions across time and cortex (\cref{sec:analysis,sec:app-analysis}).
\section{\bench: Probing Latent Brain State}
\label{sec:benchmark}

\bench evaluates brain world models across three core capabilities: inferring a latent state from neural observations, predicting its stimulus-driven evolution, and updating it as new observations arrive.
All scores are computed on the model's latent state against measured observations.

\subsection{Evaluation protocol}
\label{sec:probe}
Each evaluation window begins at an anchor step $t_0$ of a recording.
Given the individual's neural observations up to $t_0$ and the stimulus stream, the model rolls out predicted latent states over the next $T = 40$ steps (60\,s in our fMRI setup).

To compare architectures with distinct latent spaces on a common neural target~\cite{assran2023ijepa,assran2025vjepa2}, \bench evaluates predicted latent states by linearly probing how much observed neural activity (fMRI) they explain.
For each model and individual, we fit a linear probe from latent space to observation space on the training split and freeze it for all evaluations.
This protocol measures what the predicted latent state encodes about an individual's brain activity and compares all architectures on equal footing without model-specific decoders or retraining.

\subsection{Tasks and metrics}
\label{sec:tasks}
\bench defines one task and one variance-explained metric per capability (\cref{tab:metrics}).

\paragraph{Notation.}
Within each evaluation window, we index the $T = 40$ steps following $t_0$ as $t = 1, \dots, T$.
Let $x_t \in \R^D$ denote the measured observation at step $t$, $\bar x$ its temporal mean over the full recording, and $\hat x_t$ the observation read out by applying the frozen linear probe to the model's predicted latent state at step $t$.
All sums pool across $t = 1, \dots, T$ and across all anchor steps $t_0$, with scores reported as percentages ($\times 100$).

\begin{table}[t]
\centering
\small
\setlength{\tabcolsep}{4pt}
\begin{tabularx}{\linewidth}{@{}l>{\raggedright\arraybackslash}X>{\raggedright\arraybackslash}X@{}}
\toprule
Task & Model inputs & Probed target \\
\midrule
\emph{Forecast} & Observations up to $t_0$ and stimulus & Future activity $x_t$ \\
\addlinespace[4pt]
\emph{Updating} & Same as \emph{Forecast}, plus sparse new observations & Future activity $x_t$ from refreshed states \\
\addlinespace[4pt]
\emph{State} & Observations up to $t_0$ and stimulus across two trials & Trial-to-trial difference $d_t = x^{(1)}_t - x^{(2)}_t$ \\
\bottomrule
\end{tabularx}
\caption{\textbf{Overview of \bench.} Each task evaluates a model's predicted latent states via a frozen linear probe and reports variance explained (\%) in the target. \emph{Forecast} measures multi-step prediction without new observations, \emph{Updating} measures state revision as new observations arrive, and \emph{State} measures state inference across repeated trials of the same stimulus.}
\label{tab:metrics}
\end{table}

\paragraph{\emph{Forecast}: predicting future states.}
Given observations up to $t_0$ and upcoming stimulus features, the model autoregressively rolls out $T$ future latent states without access to observations after $t_0$.
We score the fraction of observation variance explained across the one-minute horizon:
\begin{equation}
\textsc{Forecast} = 1 - \frac{\sum_t \| x_t - \hat x_t \|_2^2}{\sum_t \| x_t - \bar x \|_2^2}.
\label{eq:forecast}
\end{equation}
\emph{Forecast} measures how accurately a model projects brain activity forward from its inferred initial state and the incoming stimulus.
Perfect predictions score $100\%$, predicting the recording mean $\bar x$ scores $0\%$, and predictions less accurate than $\bar x$ score below $0\%$.

\paragraph{\emph{Updating}: revising the state with new observations.}
\emph{Updating} mirrors \emph{Forecast} except that sparse observations arrive mid-rollout, mimicking online neural recording.
At four steps by default, $t \in R = \{8, 16, 24, 32\}$ (\cref{sec:app-properties} varies the budget), the model first predicts step $t$ and only then observes $x_t$ to refresh its state before continuing the rollout.
Every prediction $\hat x^{\,\mathrm{upd}}_t$ remains strictly causal (conditioning only on past observations, including at the four update steps) and is scored over the same $T$ steps:
\begin{equation}
\textsc{Updating} = 1 - \frac{\sum_t \| x_t - \hat x^{\,\mathrm{upd}}_t \|_2^2}{\sum_t \| x_t - \bar x \|_2^2}.
\end{equation}
A model that ignores incoming observations reduces to \emph{Forecast}. The gain $\textsc{Updating} - \textsc{Forecast}$ therefore directly quantifies how effectively the model incorporates new neural observations.

\paragraph{\emph{State}: inferring the state from observations.}
Open-loop forecasts blend stimulus-driven responses with history-dependent internal state.
\emph{State} isolates the latter by pairing two trials in which the same individual experiences an identical stimulus (two viewings of the same movie), yielding observations $x^{(1)}_t$ and $x^{(2)}_t$.
Conditioned on each trial's respective pre-$t_0$ history, the model produces trial-specific predictions $\hat x^{(1)}_t$ and $\hat x^{(2)}_t$.
Taking between-trial differences,
\begin{equation}
d_t = x^{(1)}_t - x^{(2)}_t, \qquad \hat d_t = \hat x^{(1)}_t - \hat x^{(2)}_t,
\end{equation}
cancels deterministic stimulus-driven activity, leaving trial-specific variability rooted in the individual's brain state.
We score the fraction of between-trial variance explained by the probed state:
\begin{equation}
\textsc{State} = 1 - \frac{\sum_t \| d_t - \hat d_t \|_2^2}{\sum_t \| d_t \|_2^2}.
\end{equation}
\emph{State} quantifies how much stimulus-unexplained neural variability the model infers from past observations.
Any stimulus-only model predicts identical trajectories across repeats ($\hat x^{(1)}_t = \hat x^{(2)}_t$, hence $\hat d_t = 0$) and scores $0\%$.

\subsection{Data}
\label{sec:data}
We instantiate \bench on the Algonauts 2025 dataset~\cite{gifford2025algonauts}, comprising whole-cortex fMRI from four individuals watching the television series \emph{Friends} and four feature films.
Signals are sampled every $1.49\,\mathrm{s}$ (one discrete step) and parcellated into $D = 1{,}000$ cortical regions~\cite{schaefer2018parcellation}.
For each individual, models are trained on \emph{Friends} Seasons 1--5 (46.6 hours) and evaluated on held-out Season 6 (9.5 hours) plus all four movies (10.2 hours).
Two held-out movies, \emph{Hidden Figures} and \emph{Life}, were viewed twice by every individual, providing the paired repeats for \emph{State}.
Evaluation windows start every 2 steps ($\approx 3\,\mathrm{s}$) for \emph{Forecast} and \emph{State}, and every 20 steps ($\approx 30\,\mathrm{s}$) for \emph{Updating}.

To validate generalization on a second dataset, we also instantiate \bench on the movie-fMRI data of the Human Connectome Project (HCP)~\cite{vanessen2013hcp}, where individuals watch short movie clips ($1\,\mathrm{s}$ per step, $D = 1{,}000$ regions), evaluating on held-out clips including a repeated clip for \emph{State} (\cref{sec:app-hcp}).
\section{Experiments}
\label{sec:experiments}

\begin{table*}[t]
\centering
\small
\setlength{\tabcolsep}{5pt}
\begin{tabularx}{\textwidth}{>{\raggedright\arraybackslash}Xcccccccccc}
\toprule
 & \multicolumn{3}{c}{7.5\,s of fMRI} & \multicolumn{3}{c}{120\,s of fMRI} & \multicolumn{3}{c}{180\,s of fMRI} & $r\uparrow$ \\
\cmidrule(lr){2-4}\cmidrule(lr){5-7}\cmidrule(lr){8-10}
Model & \emph{Forecast}$\uparrow$ & \emph{State}$\uparrow$ & \emph{Updating}$\uparrow$ & \emph{Forecast}$\uparrow$ & \emph{State}$\uparrow$ & \emph{Updating}$\uparrow$ & \emph{Forecast}$\uparrow$ & \emph{State}$\uparrow$ & \emph{Updating}$\uparrow$ & 180\,s \\
\midrule
TRIBE v2~\cite{dascoli2026tribev2} & 4.2 & 0.0 & 4.1 & 4.2 & 0.0 & 4.1 & 4.2 & 0.0 & 4.1 & 0.173 \\
MIRAGE~\cite{gokce2026mirage} & 8.8 & 0.0 & 8.8 & 8.8 & 0.0 & 8.8 & 8.8 & 0.0 & 8.8 & \underline{0.320} \\
Linear forecaster & \underline{10.4} & \textbf{1.4} & \underline{14.7} & \underline{11.6} & \underline{2.6} & \underline{14.8} & \underline{11.6} & \underline{2.6} & 12.7 & 0.294 \\
IPSID~\cite{vahidi2024ipsid} & 8.5 & 0.9 & 12.4 & 8.7 & 1.2 & 12.5 & 8.7 & 1.2 & 12.5 & 0.243 \\
NeuroWorld~\cite{dong2026neuroworld} & 8.9 & 0.2 & 13.7 & 9.2 & 0.4 & 14.4 & 8.3 & 0.4 & \underline{13.4} & 0.239 \\
BrainVista~\cite{yin2026brainvista} & 5.0 & $-$0.4 & 8.4 & 6.4 & 0.2 & 10.2 & 6.6 & 0.2 & 10.3 & 0.220 \\
\midrule
\rowcolor{rowhighlight}
\textbf{\method} & \textbf{11.5} & \underline{1.1} & \textbf{16.9} & \textbf{15.0} & \textbf{4.5} & \textbf{19.8} & \textbf{16.0} & \textbf{5.5} & \textbf{20.3} & \textbf{0.349} \\
\bottomrule
\end{tabularx}
\caption{\textbf{\method outperforms all baselines across multi-minute horizons on \bench.} Scores (\%) with 7.5\,s, 120\,s, and 180\,s of fMRI history, plus parcel-averaged 40-step Pearson correlation $r$ at 180\,s (\textbf{best}, \underline{second best}). Extending history to 180\,s boosts \method's \emph{State} score fivefold ($1.1\% \to 5.5\%$), whereas baselines plateau or decline beyond 120\,s.}
\label{tab:main}
\end{table*}

\subsection{Setup}
\label{sec:setup}

\paragraph{Baselines.}
We compare \method against six baselines across four families:
(i) \emph{TRIBE v2}~\cite{dascoli2026tribev2} and \emph{MIRAGE}~\cite{gokce2026mirage}, stimulus-only encoders without fMRI input;
(ii) a per-individual \emph{Linear forecaster} fitting independent ridge regressions per lead step from stimulus features and fMRI history;
(iii) \emph{IPSID}~\cite{vahidi2024ipsid}, a stimulus-driven linear state-space model with Kalman filtering; and
(iv) prior brain world and forecasting models: \emph{NeuroWorld}~\cite{dong2026neuroworld} (adapted with latent history tokens for multi-minute context) and \emph{BrainVista}~\cite{yin2026brainvista} (trained per individual using official code).
Latent-state models (\method, NeuroWorld, IPSID) are read out via frozen per-individual linear probes (\cref{sec:probe}), while observation-space models are scored directly on their predicted fMRI (\cref{sec:app-baselines}).

\paragraph{Implementation.}
\method trains a single shared encoder and predictor across individuals, using 180\,s of fMRI history by default, a 7.5\,s ($m = 5$) fast-state window, $1{,}024$-dimensional fast and slow states ($s_t \in \R^{2{,}048}$), a 7.5\,s stimulus window, and $K = 40$-step (60\,s) training rollouts (\cref{tab:settings}).
In \emph{Updating}, unobserved intermediate steps are imputed with the model's own forecast before encoding.

\subsection{Main results}
\label{sec:main}

Across multi-minute contexts, \method outperforms all baselines on every metric (\cref{tab:main}), for every individual (\cref{sec:app-viewers}), and across all forecast lead times and observation budgets (\cref{sec:app-properties}).
At 180\,s, it reaches $16.0\%$ on \emph{Forecast} and $20.3\%$ on \emph{Updating} ($+4.3$ and $+6.9$ points over the strongest baseline; \cref{sec:app-significance}), while more than doubling \emph{State} ($5.5\%$ vs.\ $2.6\%$), a lead that also holds for the correlation of trial-to-trial differences (\cref{sec:app-standard}).
Extending history from 7.5\,s to 180\,s boosts \method's \emph{State} fivefold ($1.1\% \to 5.5\%$), whereas baselines plateau or decline beyond 120\,s (\cref{sec:app-history}).

\method also transfers under distribution shift to the HCP data (184 individuals), with a distinct scanner, sampling rate, and stimuli (\cref{sec:app-hcp}).
Freezing the Algonauts fast encoder $g_\phi$ and fitting only the slow weights and linear predictor on 148 HCP training individuals, \method outperforms all baselines trained on HCP from scratch across all three tasks ($6.6\%$, $6.2\%$, and $15.7\%$; \cref{tab:hcp}) and on standard metrics (\cref{tab:hcp-standard}), with significant gains in most comparisons. Moreover, \method generalizes zero-shot to the remaining 36 unseen individuals: reading out its latent state with a single probe shared across the 148 training individuals reaches $7.3\%$ on \emph{Forecast} and $16.7\%$ on \emph{Updating} on their test recording without any calibration, matching accuracy on the training individuals (\cref{sec:app-newviewer}).

\paragraph{Inference efficiency.}
With 30\,M parameters, \method rolls out 60\,s in 5.4\,ms, over $20\times$ faster than Transformer-based NeuroWorld and BrainVista (\cref{sec:app-efficiency}).

\paragraph{Data efficiency.}
Retrained end-to-end on nested data subsets, with just 10\% of the episodes ($\approx 4.7$\,h per individual), \method already surpasses the full-data ($46.6$\,h) linear forecaster on \emph{State} ($3.6\%$ vs.\ $2.6\%$) and \emph{Updating} ($15.5\%$ vs.\ $12.7\%$) while matching its \emph{Forecast} accuracy ($11.5\%$ vs.\ $11.6\%$).
With 25\% ($\approx 11.7$\,h), \method surpasses the full-data forecaster across all three tasks and continues scaling up to 100\% (\cref{sec:app-datasize}).
\Cref{sec:app-generated} shows that \method's forecasts also improve other models when used as training targets.

\subsection{Beyond standard metrics}
\label{sec:inputs}

\begin{table}[t]
\centering
\small
\setlength{\tabcolsep}{3.5pt}
\begin{tabularx}{\linewidth}{>{\raggedright\arraybackslash}Xcccc}
\toprule
Input condition & \emph{Forecast} & \emph{State} & \emph{Updating} & $r$ \\
\midrule
\rowcolor{rowhighlight}
Full inputs & 16.0 & 5.5 & 20.3 & 0.349 \\
\midrule
Without stimulus & 7.5 & 5.5 & 13.7 & 0.224 \\
Without fMRI & 9.4 & 0.0 & 8.4 & 0.253 \\
Another individual's fMRI & 6.7 & $-$5.1 & 1.2 & 0.236 \\
\bottomrule
\end{tabularx}
\caption{\textbf{\bench disentangles capabilities that standard metrics conflate.} Evaluating the fitted \method (180\,s history), without refitting, when dropping its stimulus term, zeroing fMRI observations, or substituting another individual's fMRI from the same movie (\cref{sec:app-swap}). Each perturbation degrades the three scores in a distinct pattern, whereas correlation $r$ scores all three within $0.03$ of one another.}
\label{tab:inputs}
\end{table}

Prior encoding and brain world models rely on standard metrics such as parcel-wise Pearson $r$ (\cref{sec:app-standard}), which mix shared stimulus responses with individual state.
Perturbing one input of \method at a time (\cref{tab:inputs}) shows how \bench separates what these metrics conflate.
We keep the fitted \method and its probes fixed and change one input at test time: we drop the stimulus term from the forecast (\emph{without stimulus}), replace the fMRI that the model reads with zeros (\emph{without fMRI}), or feed the fMRI of another individual watching the same movie while still scoring against the original individual (\emph{another individual's fMRI}).
Both fMRI conditions apply to the history before $t_0$ and to the observations revealed in \emph{Updating} (details in \cref{sec:app-swap}).

\emph{Forecast} depends on both inputs, dropping roughly in half when the stimulus is removed ($16.0\% \to 7.5\%$) or fMRI observations are zeroed ($9.4\%$).
\emph{State} depends solely on the target individual's observations: it is unchanged without the stimulus ($5.5\%$), zero without fMRI by construction, and negative under another individual's fMRI from the same movie ($-5.1\%$).
\emph{Updating} tests how a model incorporates mid-rollout observations: its gain over \emph{Forecast} persists without the stimulus ($13.7\%$ vs.\ $7.5\%$), vanishes without fMRI ($8.4\%$ vs.\ $9.4\%$), and turns into a loss with another individual's fMRI ($1.2\%$ vs.\ $6.7\%$).

By contrast, standard metrics miss these distinctions: $r$ scores the three perturbations within $0.03$ of one another ($0.224$--$0.253$), while stimulus-only MIRAGE ranks second overall on 40-step $r$ ($0.320$, \cref{tab:main}) and global $r$, and beats every prior world model on segment identification (\cref{tab:standard}).
\bench thus diagnoses not only prediction accuracy, but which inputs a model uses and how well, confirming that both stimuli and an individual's own brain state govern naturalistic activity (\cref{sec:intro}).

\subsection{Ablations}
\label{sec:ablations}

\begin{table*}[t]
\centering
\footnotesize
\setlength{\tabcolsep}{2.2pt}
\begin{tabularx}{0.488\linewidth}[t]{@{}>{\raggedright\arraybackslash}Xcccc@{}}
\toprule
\textbf{(a) Architecture} & \emph{Forecast} & \emph{State} & \emph{Updating} & $r$ \\
\midrule
w/o slow state $u_t$ & $-4.6$ & $-4.3$ & $-3.4$ & $-0.066$ \\
Slow weights: learned $\to$ mean & $-2.8$ & $-2.8$ & $-2.6$ & $-0.040$ \\
Transformer encoder $g_\phi$ & $-1.2$ & $-1.2$ & $-1.1$ & $-0.016$ \\
Transformer predictor $p_\theta$ & $-1.2$ & $-1.1$ & $-0.7$ & $-0.015$ \\
Transformer $g_\phi$ and $p_\theta$ & $-2.7$ & $-2.1$ & $-2.4$ & $-0.032$ \\
\addlinespace[3pt]
\textbf{(b) Training} & \emph{Forecast} & \emph{State} & \emph{Updating} & $r$ \\
\midrule
w/o pretraining (random init.) & $-2.9$ & $-2.7$ & $-2.5$ & $-0.041$ \\
Frozen pretrained encoder & $-0.7$ & $-0.6$ & $+0.2$ & $-0.014$ \\
w/o EMA and stop-gradient & $-0.8$ & $-0.7$ & $+0.1$ & $-0.015$ \\
\bottomrule
\end{tabularx}\hfill
\begin{tabularx}{0.488\linewidth}[t]{@{}>{\raggedright\arraybackslash}Xcccc@{}}
\toprule
\textbf{(c) Stimulus modalities} & \emph{Forecast} & \emph{State} & \emph{Updating} & $r$ \\
\midrule
w/o video features & $-2.5$ & 0.0 & $-2.1$ & $-0.033$ \\
w/o text features & $-1.1$ & 0.0 & $-0.8$ & $-0.017$ \\
w/o audio features & $-0.6$ & 0.0 & $-0.5$ & $-0.006$ \\
\addlinespace[3pt]
\textbf{(d) Hyperparameters} & \emph{Forecast} & \emph{State} & \emph{Updating} & $r$ \\
\midrule
Fast-state window: 7.5 $\to$ 4.5\,s & $-0.1$ & $-0.1$ & $-0.1$ & $-0.002$ \\
Stimulus window: 7.5 $\to$ 1.5\,s & $-0.5$ & 0.0 & $-0.6$ & $-0.005$ \\
State size: 2,048 $\to$ 1,024 & $-0.2$ & $-0.2$ & $-0.1$ & $-0.003$ \\
State size: 2,048 $\to$ 4,096 & $+0.1$ & $+0.1$ & $+0.2$ & $+0.001$ \\
Rollout $K$: 40 $\to$ 80 & $-0.1$ & $-0.1$ & $+0.1$ & $-0.002$ \\
\bottomrule
\end{tabularx}
\caption{\textbf{Learned slow state, linear operators, and two-stage training drive \method's performance.} Changes relative to full 180\,s \method ($16.0$, $5.5$, $20.3$, $0.349$; \cref{tab:main}), in percentage points for \bench tasks and $\Delta r$ for correlation (default $\to$ ablated).}
\label{tab:ablations}
\end{table*}

In \cref{tab:ablations}, the learned slow state $u_t$ drives long-context performance: removing it drops \emph{Forecast} by $4.6$ points and \emph{State} from $5.5\%$ to $1.2\%$, reverting to 7.5\,s accuracy (\cref{tab:main}), while uniform 180\,s averaging still loses $2.8$ points.
Linear operators outperform nonlinear alternatives: replacing $g_\phi$, $p_\theta$, or both with Transformers degrades every metric ($-2.7$ combined on \emph{Forecast}) and causes rollouts to diverge beyond 2\,min, whereas \method stays stable through 5\,min (\cref{sec:app-long}).
The two training stages act at complementary horizons (\cref{tab:app-ablations}): encoder pretraining anchors near-term state inference ($-2.9$ on \emph{Forecast} without it), while latent-rollout fine-tuning ($-0.7$ when frozen) and EMA targets ($-0.8$) sustain 30--60\,s accuracy.
All three stimulus modalities help (led by video, $-2.5$), and hyperparameter changes shift scores by $\le 0.6$ points.

\subsection{Where and when state and stimulus matter}
\label{sec:analysis}

\begin{figure}[t]
\centering
\includegraphics[width=\linewidth]{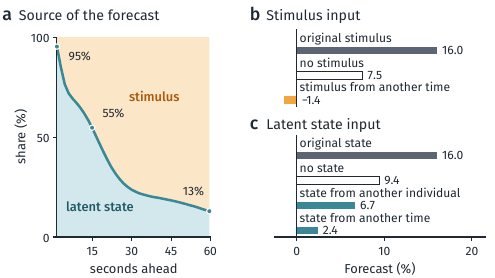}
\caption{\textbf{Latent state governs near-term forecasts while the stimulus dominates longer horizons.} Decomposing 180\,s \method rollouts on held-out Algonauts recordings (\cref{sec:app-analysis}): (a) variance share of the latent-state versus stimulus component over the 60\,s rollout; (b, c) \emph{Forecast} accuracy when omitting a component or swapping it across time or individuals.}
\label{fig:time}
\end{figure}

\begin{figure*}[t]
\centering
\includegraphics[width=\linewidth]{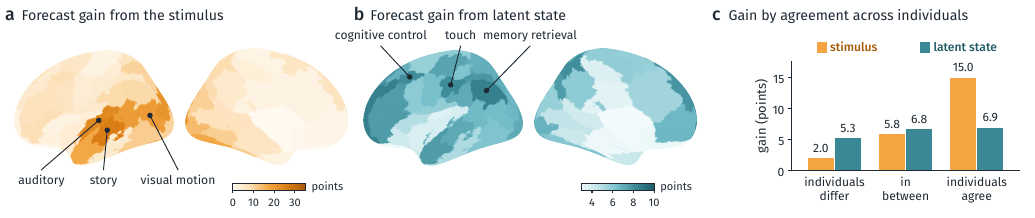}
\caption{\textbf{Stimuli drive shared sensory and narrative-processing cortex, whereas latent state improves prediction cortex-wide.} \emph{Forecast} gain from adding (a) the stimulus to state-only and (b) the latent state to stimulus-only predictions on the left hemisphere (100 Schaefer regions~\cite{schaefer2018parcellation}; top lateral regions labeled via NeuroQuery~\cite{dockes2020neuroquery}). (c) Mean gain across 1,000-parcel tertiles of inter-subject correlation.}
\label{fig:cortex}
\end{figure*}

Because \method's linear rollout decomposes each forecast into exact initial-state and stimulus terms without retraining (\cref{sec:linear,sec:app-analysis}), we can map how ongoing internal dynamics and incoming stimuli jointly organize cortical activity on held-out Algonauts recordings.
Together, they reveal a clear spatiotemporal division of labor: latent state governs near-term dynamics and individual-specific association cortex, whereas the stimulus drives longer horizons in shared sensory and narrative-processing regions.

\paragraph{Across time.}
The initial-state component dominates early forecasts, with its relative variance share falling from 95\% at 1.5\,s to 13\% at 60\,s (\cref{fig:time}a), supported by slow latent eigenmodes persisting beyond 30\,s (\cref{sec:app-dynamics}).
Crucially, substituting another individual's state at the same movie moment ($6.7\%$ vs.\ $9.4\%$ without state), the same individual's state from another viewing of a repeated movie ($4.5\%$ vs.\ $7.4\%$), or inputs from another time ($-1.4\%$ stimulus, $2.4\%$ state) all score below omitting the component (\cref{fig:time}b,c; \cref{sec:app-swap}), confirming that latent state tracks each individual's trial- and moment-specific trajectory.

\paragraph{Across the cortex.}
Stimulus gains concentrate in auditory, visual-motion, and temporoparietal narrative-processing regions (\cref{fig:cortex}a), closely tracking parcel inter-subject correlation~\cite{hasson2004intersubject} ($r = 0.92$) where individuals respond alike.
By contrast, latent-state gains are positive in \emph{every} parcel, peak in prefrontal and parietal association cortex (\cref{fig:cortex}b; driven primarily by default-mode and control hubs, \cref{fig:relies}), and hold steady at $5$--$7$ points even where individuals diverge, exceeding stimulus gains in $59\%$ of all 1,000 parcels (\cref{fig:cortex}c; \cref{sec:app-cortex}).
\method thus separates shared stimulus-locked responses from individual-specific cortical dynamics~\cite{hasson2004intersubject,fox2006coherent}.

\section{Discussion}
\label{sec:discussion}

\paragraph{Evaluating world models beyond prediction accuracy.}
World models are increasingly used in vision, medicine, and neuroscience for systems shaped by both internal state and external inputs, yet standard evaluations still rely on output prediction accuracy~\cite{shang2026worldarena,zhang2026worldinworld}.
When external inputs strongly drive the signal, a model can score well simply by predicting the shared response to the input without tracking internal state at all (\cref{sec:inputs}).
\bench addresses this with two simple tests that apply beyond neuroimaging: comparing repeated trials of the same input to isolate what the latent state captures (\emph{State}), and providing new observations during a rollout to test whether the model updates its predictions (\emph{Updating}).

\paragraph{Two-timescale latent prediction for interpretable world models.}
\method's gains come from how it represents and evolves state over time rather than from complex nonlinear architectures (\cref{sec:main,sec:ablations}).
First, predicting in a shared latent space with a multi-step joint-embedding objective avoids fitting separate regressions at each future step to noisy fMRI, allowing strong generalization from under 5 hours instead of 47 hours of recordings per individual (\cref{sec:main,sec:app-datasize}).
Second, splitting the state into fast and slow channels with learned multi-minute weights captures slow internal context where simply feeding longer histories into a Transformer plateaus.
Third, combining rich pretrained stimulus features with linear latent dynamics outperforms deeper Transformer encoders and predictors (\cref{sec:ablations}), keeps five-minute rollouts stable (\cref{sec:app-long}), and cleanly separates each forecast into exact latent-state and stimulus terms (\cref{sec:linear,sec:analysis}).

\paragraph{Individual brain states in a shared latent space.}
Stimulus-only encoding models predict an individual's response from the stimulus, without inferring their ongoing state from neural observations~\cite{dascoli2026tribev2,gokce2026mirage,gifford2025algonauts}.
Our results reveal a predictable, multi-minute brain state that improves prediction in every cortical parcel, especially in prefrontal and parietal cortex, and is anchored in default-mode and control regions (\cref{sec:analysis}). Default-mode regions integrate incoming narratives with an individual's prior context~\cite{yeshurun2021dmn}.
Two individuals' inferred states cannot be swapped at the same movie moment (\cref{sec:inputs,sec:analysis}), yet a single shared encoder and predictor track each individual's own state trajectory, and the fast encoder trained on four Algonauts individuals transfers across scanners and stimuli to 184 HCP individuals (\cref{sec:setup,sec:main,sec:app-hcp,sec:app-newviewer}).
Different individuals thus follow personal state trajectories that can be modeled within a common latent space.

\section{Conclusion}
\label{sec:conclusion}

\looseness=1
Calling a model a world model makes a functional claim about its latent state: that it infers internal context from observations, evolves it under incoming inputs, and revises it as new measurements arrive.
\bench tests these capabilities directly on whole-cortex fMRI, and \method shows that two-timescale linear dynamics over multi-minute histories track individual brain state accurately and efficiently across datasets.
By separating trial-specific state in association cortex from shared stimulus responses, this framework provides a foundation for linking inferred brain state to behavior and \mbox{closed-loop stimulation}.

{
    \small
    \bibliographystyle{ieeenat_fullname}
    \bibliography{main}
}

\clearpage
\endgroup

\onecolumn
\appendix
\raggedbottom
\clubpenalty=10000
\widowpenalty=10000
\leftlinenumbers
\setlength{\linenumbersep}{4pt}

\captionsetup{hypcap=false}
\newenvironment{appendixblock}[1][6pt]{\par\addvspace{#1}\nolinenumbers\noindent\begin{minipage}{\textwidth}}{\end{minipage}\par\linenumbers\medskip}

\begin{appendixblock}[0pt]
\section{Implementation details}
\label{sec:app-implementation}

This section details the model configuration, baseline implementations, and computational costs supporting \cref{sec:setup,sec:main}.

\subsection{Model and training settings}
\label{sec:app-settings}


\Cref{tab:settings} lists the architecture and training hyperparameters of \method.

\medskip
\begingroup
\centering
\small
\renewcommand{\arraystretch}{0.92}
\begin{tabularx}{\linewidth}{@{}>{\raggedright\arraybackslash}X>{\raggedright\arraybackslash}X@{}}
\toprule
Setting & Value \\
\midrule
\multicolumn{2}{@{}l}{\textit{Latent state}} \\
Observation history & 120 steps (180\,s of fMRI) \\
Fast-state window, $m$ & 5 steps (7.5\,s) \\
Slow-state lags, $L$ & 116 \\
Dimensions of $f_t$ and $u_t$ & 1,024 each ($s_t \in \R^{2{,}048}$) \\
\addlinespace
\multicolumn{2}{@{}l}{\textit{Encoder pretraining}} \\
Pretraining predictor & 6-block Transformer \\
Optimizer, learning rate, batch & AdamW, $5 \times 10^{-5}$, 1,024 \\
Epochs (best on Friends S6 kept) & 100 \\
Target of $h_\phi$ & Mean of next 16 fast states (24\,s) \\
\addlinespace
\multicolumn{2}{@{}l}{\textit{Latent prediction}} \\
Learning rate, encoder / predictor & $10^{-5}$ / $5 \times 10^{-5}$ \\
Optimizer, batch & AdamW, 1,024 \\
Rollout horizon, $K$ & 40 steps (60\,s) \\
EMA target momentum & 0.998 \\
Stimulus input & 1 step, shifted 2 steps (3\,s) earlier \\
Stimulus dropout & 30\% of sequences \\
\addlinespace
\multicolumn{2}{@{}l}{\textit{Final linear predictor}} \\
Stimulus window & Last 5 steps (7.5\,s) \\
Base ridge penalty, $\lambda$ & $10^{-3} \times$ mean Gram diagonal \\
Penalty multiplier, state columns & $10\times$ \\
Penalty multiplier, stimulus columns & $100\times$ ($10{,}000\times$ on HCP) \\
Stimulus normalization & Standardized on training steps \\
\bottomrule
\end{tabularx}
\captionof{table}{\textbf{Hyperparameters and training settings of \method} across model components and training stages.}
\label{tab:settings}
\par
\endgroup
\end{appendixblock}

During latent-prediction training, the predictor conditions on a single stimulus step (3\,s earlier) and drops the stimulus in 30\% of sequences so rollouts rely on the inferred latent state.
When fitting the closed-form ridge predictor (\cref{sec:linear}), we condition on the last five stimulus steps (7.5\,s) and scale $\lambda$ separately on the state ($10\times$) and stimulus ($100\times$; $10{,}000\times$ on HCP) columns of $W$.

\begin{appendixblock}
\subsection{Baseline settings}
\label{sec:app-baselines}

\Cref{tab:baselines} summarizes how each baseline in \cref{sec:setup} is trained, conditioned, and scored on Algonauts (\cref{sec:app-hcp} details HCP adaptations).
All models use the same training split (Friends Seasons 1--5), held-out recordings, anchor steps, and reference baselines as \method (\cref{sec:probe,tab:settings}).
For \cref{fig:history}, every fMRI-reading baseline except IPSID is retrained or refitted from scratch at each history length $L$; IPSID's state-space parameters do not depend on $L$, so its Kalman filter runs over the last $L$\,s at test time.
\method, the linear forecaster, and IPSID share the same standardized 2,048-dimensional TRIBE~v2 causal contextual stimulus features.

\medskip
\begingroup
\centering
\footnotesize
\setlength{\tabcolsep}{4pt}
\begin{tabularx}{\linewidth}{@{}l>{\raggedright\arraybackslash\hsize=1.14\hsize}X>{\raggedright\arraybackslash\hsize=1.02\hsize}X>{\raggedright\arraybackslash\hsize=1.08\hsize}X>{\raggedright\arraybackslash\hsize=0.90\hsize}X>{\raggedright\arraybackslash\hsize=0.86\hsize}X@{}}
\toprule
Model & Training & fMRI history & Stimulus input & Scoring readout & Unrevealed steps (\emph{Updating}) \\
\midrule
TRIBE~v2~\cite{dascoli2026tribev2} & Frozen pretrained & None & Built-in video, audio, text & Average-individual head & Not used \\
MIRAGE~\cite{gokce2026mirage} & Frozen pretrained & None & Qwen3-Omni~\cite{xu2025qwen3omni} & Per-individual heads & Not used \\
Linear forecaster & Closed-form ridge (per individual, per $L$) & Per-step ($\le 30$\,s); 15\,s means ($>30$\,s) & TRIBE~v2 contextual (lags 3--8) & Direct fMRI output & Own forecast \\
IPSID~\cite{vahidi2024ipsid} & Subspace ID (per individual) & Kalman filter over $L$\,s & TRIBE~v2 contextual (32 PCs, lag 3) & Observation matrix & Skip update \\
NeuroWorld~\cite{dong2026neuroworld} & End-to-end latent + decoder (shared, per $L$) & 3 latest steps + 7.5\,s tokens & TRIBE~v2 modality projections & Frozen linear probe & Decoded forecast \\
BrainVista~\cite{yin2026brainvista} & Autoencoder + predictor (per individual, per $L$) & Every step over $L$\,s & V-JEPA~2, w2v-BERT~2.0, Llama~3.2 (384 PCs each) & Rescaled fMRI output & Own forecast \\
\bottomrule
\end{tabularx}
\captionof{table}{\textbf{Summary of baseline configurations on Algonauts.} \emph{Scoring readout} specifies how model outputs or states map to parcel predictions, and the final column specifies how unrevealed intermediate steps are handled during \emph{Updating}. HCP adaptations are described in \cref{sec:app-hcp}.}
\label{tab:baselines}
\par
\endgroup
\end{appendixblock}

\paragraph{TRIBE v2.}
TRIBE~v2 processes consecutive 100\,s segments of 2\,Hz multimodal features with bidirectional attention and predicts fsaverage5 surface activity via its average-individual head (including a built-in 5\,s hemodynamic offset).
Vertex predictions within 3\,mm of the mid-thickness surface are averaged into the 1,000 Schaefer parcels~\cite{schaefer2018parcellation} and temporally pooled onto the fMRI grid.

\paragraph{MIRAGE.}
MIRAGE maps video, audio, and transcript features from Qwen3-Omni-30B-A3B-Thinking~\cite{xu2025qwen3omni} to each individual's cortical activity through released per-individual heads.
Output steps align directly with fMRI steps at zero shift, where correlation with measured fMRI peaks.

\paragraph{Linear forecaster.}
For each individual and history length $L$, stage one fits a closed-form ridge regression from stimulus lags 3--8 (4.5--11.9\,s) to fMRI ($\alpha \in \{10^3,\dots,10^7\}$ chosen by Season~6 Pearson $r$).
Stage two fits a separate ridge regression at each lead step $k=1,\dots,40$ ($\beta_k \in \{10^1,\dots,10^8\}$ chosen by Season~6 normalized squared error) to predict stage-one residuals from whitened spatial PCs of fMRI history (within a 2,048-feature budget: up to 256 PCs/step for $L \le 30$\,s, and 102 PCs/step over the last 30\,s plus 15\,s block means for older history).
Training windows start every 2 steps and require complete fMRI over the last $\min(L, 30\,\mathrm{s})$, filling older missing steps with the training mean.

\paragraph{IPSID.}
Using the authors' PSID package ($n_1=0$, 10 block rows) on every second training episode, we fit one input-driven linear state-space model per individual over the top 64 spatial fMRI PCs and top 32 stimulus PCs (3-step, 4.5\,s lag).
The state dimension is selected from $\{16, 32\}$ by 20-step Season~6 error (yielding 32, 16, 32, and 32 states for individuals 1, 2, 3, and 5).
At evaluation, a zero-initialized steady-state Kalman filter processes the last $L$\,s of history before rolling out open loop under the stimulus and projecting back from PC space to parcels.

\paragraph{NeuroWorld.}
Following~\cite{dong2026neuroworld} (no public code), a shared linear encoder ($1{,}024$-D) and 6-block Transformer predictor (width 1,024, 16 heads, MLP width 2,048) predict the next state from the 3 latest states (our 7.5\,s row), conditioned on TRIBE~v2 modality projections (5\,s lag) via adaptive layer normalization; longer contexts prepend 7.5\,s (5-step mean) history tokens (15 tokens at 120\,s; 23 at 180\,s).
Stage one trains end to end with AdamW (lr $5\times 10^{-5}$, batch 1,024, up to 100 epochs with Season~6 early stopping) on next-step plus 16-step rollout MSE regularized by SIGReg (weight 0.09)~\cite{maes2026lewm}.
Stage two trains per-individual MLP decoders (width 1,024; lr $10^{-5}$, 200 epochs, selected by Season~6 Pearson $r$) on frozen 20-step rollouts; decoders are used only to impute and re-encode unrevealed steps during \emph{Updating}.

\paragraph{BrainVista.}
Using the official code on per-individual z-scored parcels, seven 2-layer MLP autoencoders (width 768; AdamW, lr $3\times 10^{-4}$, 40 epochs, selected on Season~6) compress the seven Yeo networks~\cite{yeo2011organization} into tokens of size $\approx 0.8\times$ each network's parcel count (capturing 85--94\% of Season~6 variance).
A 4-layer causal predictor (width 256, 8 heads, S2B mask and spatial mixer; context $6$, $81$, or $121$ steps) predicts tokens one step ahead from past tokens and 384-PC features from V-JEPA~2~\cite{assran2025vjepa2}, w2v-BERT~2.0~\cite{barrault2023seamless}, and Llama~3.2-3B~\cite{grattafiori2024llama3} (AdamW, lr $2\times 10^{-4}$, batch 32, 40 epochs, selected by 20-step Season~6 $r$).
To retain long-history windows, only the last $\min(L, 30\,\mathrm{s})$ requires complete fMRI, with older missing steps filled by zero-valued step tokens.

\paragraph{Training compute.}
NeuroWorld trained on 1--4 A100 GPUs, BrainVista predictors took 46--135\,min per run on one V100-32GB GPU, MIRAGE feature extraction ran on A100-80GB GPUs, and both the linear forecaster and IPSID fitted on CPUs within minutes per individual.


\begin{appendixblock}
\subsection{Model size and inference cost}
\label{sec:app-efficiency}

\Cref{tab:efficiency} compares parameter counts, multiply-add FLOPs, and single-GPU inference latency for one 60\,s forecast.

\medskip
\begingroup
\centering
\small
\begin{tabularx}{\linewidth}{>{\raggedright\arraybackslash}X*{3}{>{\centering\arraybackslash}X}}
\toprule
 & Parameters$\downarrow$ & FLOPs$\downarrow$ & Latency$\downarrow$ \\
Model & (M) & (G) & (ms) \\
\midrule
TRIBE v2~\cite{dascoli2026tribev2} & 137.3 & 16.6 & 5.3 \\
MIRAGE~\cite{gokce2026mirage} & 1{,}123.1 & 348.8 & 44.7 \\
Linear forecaster & 539.1 & 0.6 & 1.2 \\
IPSID~\cite{vahidi2024ipsid} & 0.2 & 0.02 & 17.0 \\
NeuroWorld~\cite{dong2026neuroworld} & 91.5 & 97.7 & 179.3 \\
BrainVista~\cite{yin2026brainvista} & 28.5 & 38.0 & 130.4 \\
\midrule
\rowcolor{rowhighlight}
\textbf{\method} & 30.4 & 1.7 & 5.4 \\
\bottomrule
\end{tabularx}
\captionof{table}{\textbf{\method computes rollouts over $20\times$ faster than Transformer-based fMRI world models.} Post-feature parameter count (summed across four individuals for per-individual models), multiply-add FLOPs, and single-A100 latency (batch size 1) for one 60\,s (40-step) forecast from 180\,s history. Costs for longer-window encoders (TRIBE v2, MIRAGE) are normalized per 60\,s.}
\label{tab:efficiency}
\par
\endgroup
\end{appendixblock}

Because \method's rollout applies matrix products at each step rather than evaluating a Transformer across all 40 steps, it requires over $20\times$ fewer FLOPs and runs over $20\times$ faster than NeuroWorld and BrainVista, while matching BrainVista's parameter count and using one third of NeuroWorld's.
IPSID's low-dimensional PCA state is lightweight but sacrifices nearly half of \method's forecast accuracy (\cref{tab:main}), whereas fitting separate regressions per lead step and per individual gives the linear forecaster $18\times$ as many parameters as \method.

\begin{appendixblock}
\section{Additional benchmark results}
\label{sec:app-results}

We expand \cref{sec:main,sec:inputs} across fMRI history lengths, forecast horizons, individuals, standard metrics, and transfer to the Human Connectome Project.

\subsection{Scaling with history length}
\label{sec:app-history}

\Cref{fig:history} extends \cref{tab:main} across fMRI history lengths from 7.5\,s to 180\,s.

\medskip
\begingroup
\centering
\includegraphics[width=\linewidth]{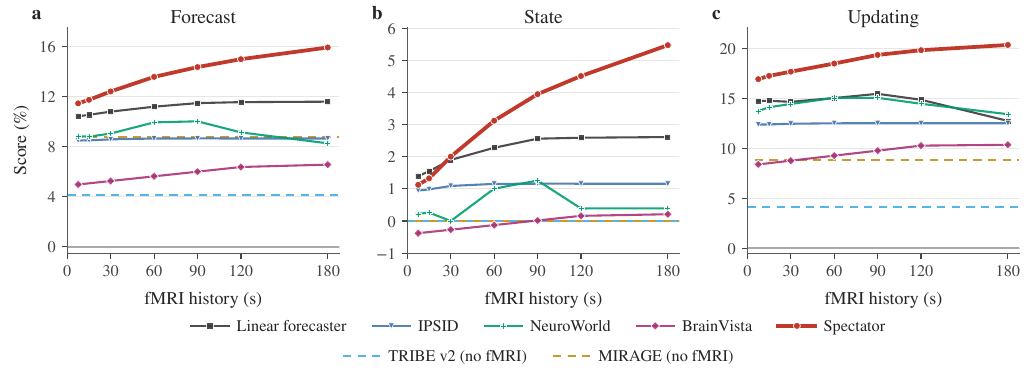}
\captionof{figure}{\textbf{\method uniquely scales with multi-minute neural history.} Performance on (a) \emph{Forecast}, (b) \emph{State}, and (c) \emph{Updating} across fMRI history lengths from 7.5\,s to 180\,s. Baselines plateau or degrade at longer contexts, whereas \method improves steadily across all three tasks (dashed lines: stimulus-only models without fMRI input).}
\label{fig:history}
\par
\endgroup
\end{appendixblock}

\method's \emph{Forecast} accuracy rises monotonically with context length, while \emph{State} grows fivefold from $1.1\%$ at 7.5\,s to $5.5\%$ at 180\,s.
By contrast, baselines fail to exploit multi-minute context: the linear forecaster plateaus on \emph{Forecast} and \emph{State} beyond 120\,s while dropping from $14.8\%$ to $12.7\%$ on \emph{Updating}; IPSID barely changes; NeuroWorld peaks at 60--90\,s before declining; and BrainVista improves on \emph{Forecast} and \emph{Updating} while remaining near zero on \emph{State}.
\method's margin over every baseline therefore widens with history length across all three tasks.

\begin{appendixblock}
\subsection{Horizon decay and observation budgets}
\label{sec:app-properties}

\Cref{fig:properties} resolves \bench performance by forecast lead time and intermediate observation budget.

\medskip
\begingroup
\centering
\includegraphics[width=\linewidth]{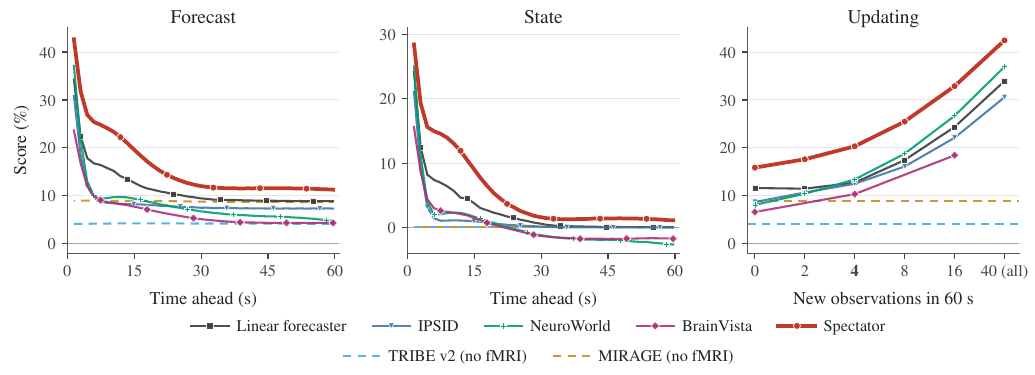}
\captionof{figure}{\textbf{\method leads at every forecast horizon and observation budget} (180\,s fMRI history). \emph{Left, Middle:} \emph{Forecast} and \emph{State} across lead times, showing informative state predictions well beyond initial steps. \emph{Right:} \emph{Updating} versus number of intermediate observations supplied during the 60\,s rollout (default: 4, bold).}
\label{fig:properties}
\par
\endgroup
\end{appendixblock}

For fMRI-conditioned models, \emph{Forecast} and \emph{State} peak immediately after the anchor step and decay with lead time (\cref{fig:properties}, left and middle).
For \method, \emph{Forecast} declines from $42.7\%$ at 1.5\,s to $11.3\%$ at 60\,s, while \emph{State} averages $18.6\%$ over 0--7.5\,s and $6.5\%$ over 7.5--30\,s, confirming that the inferred state remains informative well beyond initial steps.
On \emph{Updating} (\cref{fig:properties}, right), accuracy rises steadily with each revealed observation and reduces to single-step prediction when all 40 steps are supplied.
Each revealed observation immediately sharpens fast-timescale steps ($+26.7$ points on the next step with 2 observations), while earlier updates compound in the slow state: measured 4.5\,s after the final revealed step, the gain grows from $+2.5$ points with 2 observations to $+7.2$ points with 16.

\begin{appendixblock}
\subsection{Statistical significance}
\label{sec:app-significance}


\Cref{tab:significance} reports paired bootstrap differences between \method and each baseline across all entries of \cref{tab:main}.

\medskip
\begingroup
\centering
\small
\begin{tabularx}{\linewidth}{@{}l c >{\centering\arraybackslash\hsize=0.90\hsize}X >{\centering\arraybackslash\hsize=0.90\hsize}X >{\centering\arraybackslash\hsize=0.96\hsize}X >{\centering\arraybackslash\hsize=1.24\hsize}X@{}}
\toprule
Baseline & fMRI & \emph{Forecast} & \emph{State} & \emph{Updating} & $r$ \\
\midrule
TRIBE v2~\cite{dascoli2026tribev2} & 7.5\,s & $+$7.3 [6.8, 7.7] & $+$1.1 [1.0, 1.2] & $+$12.8 [12.3, 13.1] & $+$0.111 [0.107, 0.115] \\
 & 120\,s & $+$10.8 [10.5, 11.2] & $+$4.5 [4.1, 4.9] & $+$15.6 [15.3, 16.0] & $+$0.164 [0.161, 0.167] \\
 & 180\,s & $+$11.8 [11.4, 12.1] & $+$5.5 [5.0, 5.9] & $+$16.2 [15.8, 16.6] & $+$0.176 [0.173, 0.179] \\
\addlinespace
MIRAGE~\cite{gokce2026mirage} & 7.5\,s & $+$2.7 [1.3, 3.9] & $+$1.1 [1.0, 1.2] & $+$8.1 [6.7, 9.3] & $-$0.036 [$-$0.048, $-$0.025] \\
 & 120\,s & $+$6.2 [4.9, 7.4] & $+$4.5 [4.1, 4.9] & $+$11.0 [9.6, 12.2] & $+$0.017 [0.006, 0.027] \\
 & 180\,s & $+$7.2 [5.9, 8.4] & $+$5.5 [5.0, 5.9] & $+$11.5 [10.1, 12.7] & $+$0.029 [0.018, 0.039] \\
\addlinespace
Linear forecaster & 7.5\,s & $+$1.0 [1.0, 1.1] & $-$0.3 [$-$0.3, $-$0.3] & $+$2.2 [2.1, 2.3] & $+$0.014 [0.013, 0.015] \\
 & 120\,s & $+$3.4 [3.4, 3.5] & $+$1.9 [1.8, 2.1] & $+$5.0 [4.8, 5.1] & $+$0.046 [0.044, 0.047] \\
 & 180\,s & $+$4.3 [4.2, 4.4] & $+$2.9 [2.7, 3.0] & $+$7.6 [7.5, 7.7] & $+$0.055 [0.054, 0.057] \\
\addlinespace
IPSID~\cite{vahidi2024ipsid} & 7.5\,s & $+$3.0 [2.8, 3.2] & $+$0.2 [0.1, 0.2] & $+$4.5 [4.3, 4.7] & $+$0.044 [0.042, 0.046] \\
 & 120\,s & $+$6.3 [6.2, 6.5] & $+$3.3 [3.1, 3.6] & $+$7.3 [7.1, 7.5] & $+$0.093 [0.092, 0.095] \\
 & 180\,s & $+$7.3 [7.1, 7.4] & $+$4.3 [4.0, 4.6] & $+$7.8 [7.6, 8.0] & $+$0.106 [0.104, 0.108] \\
\addlinespace
NeuroWorld~\cite{dong2026neuroworld} & 7.5\,s & $+$2.6 [2.4, 2.9] & $+$0.9 [0.8, 1.0] & $+$3.2 [3.0, 3.4] & $+$0.033 [0.031, 0.035] \\
 & 120\,s & $+$5.9 [5.7, 6.0] & $+$4.1 [3.9, 4.3] & $+$5.4 [5.2, 5.5] & $+$0.079 [0.077, 0.081] \\
 & 180\,s & $+$7.7 [7.5, 7.8] & $+$5.1 [4.8, 5.3] & $+$6.9 [6.8, 7.1] & $+$0.110 [0.107, 0.112] \\
\addlinespace
BrainVista~\cite{yin2026brainvista} & 7.5\,s & $+$6.5 [6.3, 6.7] & $+$1.5 [1.4, 1.6] & $+$8.5 [8.3, 8.7] & $+$0.100 [0.096, 0.105] \\
 & 120\,s & $+$8.6 [8.4, 8.8] & $+$4.3 [4.2, 4.5] & $+$9.5 [9.3, 9.8] & $+$0.124 [0.120, 0.128] \\
 & 180\,s & $+$9.4 [9.1, 9.6] & $+$5.3 [5.0, 5.5] & $+$10.0 [9.8, 10.2] & $+$0.129 [0.125, 0.134] \\
\bottomrule
\end{tabularx}
\captionof{table}{\textbf{\method's gains at 120\,s and 180\,s are statistically significant across all baselines and metrics.} Paired differences (\method minus baseline, in percentage points for \bench tasks and $\Delta r$ for correlation) with 95\% confidence intervals from 2,000 paired bootstrap resamples over held-out recordings.}
\label{tab:significance}
\par
\endgroup
\end{appendixblock}

Every 95\% confidence interval excludes zero: \method outperforms all baselines significantly across every metric at 120\,s and 180\,s, and at 7.5\,s in all comparisons except against the linear forecaster on \emph{State} and MIRAGE on $r$.

\begin{appendixblock}
\subsection{Per-individual results}
\label{sec:app-viewers}


\Cref{tab:viewers} breaks down the 180\,s results of \cref{tab:main} by individual.

\medskip
\begingroup
\centering
\small
\setlength{\tabcolsep}{3.8pt}
\begin{tabularx}{\textwidth}{>{\raggedright\arraybackslash}X*{16}{c}}
\toprule
 & \multicolumn{4}{c}{\emph{Forecast}$\uparrow$} & \multicolumn{4}{c}{\emph{State}$\uparrow$} & \multicolumn{4}{c}{\emph{Updating}$\uparrow$} & \multicolumn{4}{c}{$r\uparrow$} \\
\cmidrule(lr){2-5}\cmidrule(lr){6-9}\cmidrule(lr){10-13}\cmidrule(lr){14-17}
Model & 1 & 2 & 3 & 5 & 1 & 2 & 3 & 5 & 1 & 2 & 3 & 5 & 1 & 2 & 3 & 5 \\
\midrule
TRIBE v2~\cite{dascoli2026tribev2} & 3.9 & 4.0 & 5.0 & 3.7 & 0.0 & 0.0 & 0.0 & 0.0 & 3.9 & 4.0 & 5.0 & 3.6 & 0.172 & 0.172 & 0.181 & 0.167 \\
MIRAGE~\cite{gokce2026mirage} & 7.8 & 7.7 & 10.9 & 8.4 & 0.0 & 0.0 & 0.0 & 0.0 & 7.9 & 7.7 & 11.0 & 8.4 & \underline{0.326} & \underline{0.316} & \underline{0.338} & \underline{0.300} \\
Linear forecaster & \underline{11.7} & \underline{11.0} & \underline{13.0} & \underline{10.5} & \underline{2.8} & \underline{2.0} & \underline{2.9} & \underline{2.7} & 13.0 & 11.5 & 14.1 & 12.0 & 0.297 & 0.286 & 0.314 & 0.279 \\
IPSID~\cite{vahidi2024ipsid} & 9.1 & 8.3 & 9.6 & 7.6 & 1.4 & 0.7 & 1.4 & 1.2 & 13.1 & 11.2 & 13.8 & 11.6 & 0.252 & 0.234 & 0.260 & 0.228 \\
NeuroWorld~\cite{dong2026neuroworld} & 8.6 & 8.5 & 8.8 & 7.2 & 0.6 & 0.5 & $-$0.1 & 0.5 & \underline{14.3} & \underline{12.6} & \underline{14.1} & \underline{12.4} & 0.247 & 0.239 & 0.247 & 0.224 \\
BrainVista~\cite{yin2026brainvista} & 6.3 & 6.2 & 8.0 & 5.7 & 0.1 & $-$0.9 & $-$0.3 & 1.8 & 10.3 & 9.7 & 11.9 & 9.2 & 0.213 & 0.203 & 0.238 & 0.226 \\
\midrule
\rowcolor{rowhighlight}
\textbf{\method} & \textbf{16.3} & \textbf{15.5} & \textbf{17.1} & \textbf{14.8} & \textbf{5.9} & \textbf{4.8} & \textbf{5.5} & \textbf{5.6} & \textbf{20.9} & \textbf{18.9} & \textbf{21.8} & \textbf{19.4} & \textbf{0.353} & \textbf{0.341} & \textbf{0.365} & \textbf{0.337} \\
\bottomrule
\end{tabularx}
\captionof{table}{\textbf{\method outperforms all baselines consistently for every individual.} Scores for individuals 1, 2, 3, and 5 (180\,s fMRI history) in \% for \bench tasks and Pearson $r$ (\textbf{best}, \underline{second best}).}
\label{tab:viewers}
\par
\endgroup
\end{appendixblock}

\method ranks first for every individual across all three \bench tasks and Pearson $r$, exceeding the linear forecaster by more than $4$ points on \emph{Forecast} while roughly doubling \emph{State}, and leading second-place NeuroWorld by more than $6$ points on \emph{Updating}.

\begin{appendixblock}
\subsection{Standard metrics}
\label{sec:app-standard}


\Cref{tab:standard} evaluates the models of \cref{tab:main} on the standard metrics used by prior brain world models~\cite{dong2026neuroworld} and encoding models, scored on the 40-step rollouts of \emph{Forecast}.

\medskip
\begingroup
\centering
\small
\setlength{\tabcolsep}{4pt}
\begin{tabularx}{\linewidth}{>{\raggedright\arraybackslash\hsize=1.4\hsize}X*{5}{>{\centering\arraybackslash\hsize=0.92\hsize}X}}
\toprule
 & \multicolumn{2}{c}{Parcel $r\uparrow$} & \multicolumn{2}{c}{Global $r\uparrow$} & Cls@Top10$\uparrow$ \\
\cmidrule(lr){2-3} \cmidrule(lr){4-5}
Model & 1--20 & 1--40 & 1--20 & 1--40 & (\%) \\
\midrule
\multicolumn{6}{l}{\emph{7.5\,s of fMRI}} \\
TRIBE v2~\cite{dascoli2026tribev2} & 0.173 & 0.173 & 0.217 & 0.217 & 25.5 \\
MIRAGE~\cite{gokce2026mirage} & \textbf{0.321} & \textbf{0.320} & \textbf{0.366} & \textbf{0.365} & \underline{68.2} \\
Linear forecaster & 0.293 & 0.270 & 0.333 & 0.313 & 66.3 \\
IPSID~\cite{vahidi2024ipsid} & 0.258 & 0.240 & 0.300 & 0.283 & 44.1 \\
NeuroWorld~\cite{dong2026neuroworld} & 0.272 & 0.251 & 0.309 & 0.291 & 50.4 \\
BrainVista~\cite{yin2026brainvista} & 0.210 & 0.184 & 0.238 & 0.216 & 26.7 \\
\rowcolor{rowhighlight}
\textbf{\method} & \underline{0.304} & \underline{0.284} & \underline{0.347} & \underline{0.328} & \textbf{69.7} \\
\addlinespace
\multicolumn{6}{l}{\emph{120\,s of fMRI}} \\
TRIBE v2~\cite{dascoli2026tribev2} & 0.173 & 0.173 & 0.217 & 0.217 & 25.5 \\
MIRAGE~\cite{gokce2026mirage} & 0.321 & \underline{0.320} & 0.366 & \underline{0.365} & 68.2 \\
Linear forecaster & \underline{0.334} & 0.291 & \underline{0.371} & 0.332 & \underline{81.2} \\
IPSID~\cite{vahidi2024ipsid} & 0.266 & 0.243 & 0.306 & 0.286 & 46.1 \\
NeuroWorld~\cite{dong2026neuroworld} & 0.305 & 0.258 & 0.342 & 0.297 & 60.5 \\
BrainVista~\cite{yin2026brainvista} & 0.263 & 0.213 & 0.286 & 0.242 & 39.5 \\
\rowcolor{rowhighlight}
\textbf{\method} & \textbf{0.395} & \textbf{0.337} & \textbf{0.427} & \textbf{0.376} & \textbf{92.2} \\
\addlinespace
\multicolumn{6}{l}{\emph{180\,s of fMRI}} \\
TRIBE v2~\cite{dascoli2026tribev2} & 0.173 & 0.173 & 0.217 & 0.217 & 25.5 \\
MIRAGE~\cite{gokce2026mirage} & 0.321 & \underline{0.320} & 0.366 & \underline{0.365} & 68.2 \\
Linear forecaster & \underline{0.340} & 0.294 & \underline{0.377} & 0.335 & \underline{84.6} \\
IPSID~\cite{vahidi2024ipsid} & 0.266 & 0.243 & 0.306 & 0.286 & 46.1 \\
NeuroWorld~\cite{dong2026neuroworld} & 0.281 & 0.239 & 0.319 & 0.280 & 52.6 \\
BrainVista~\cite{yin2026brainvista} & 0.268 & 0.220 & 0.287 & 0.245 & 42.2 \\
\rowcolor{rowhighlight}
\textbf{\method} & \textbf{0.408} & \textbf{0.349} & \textbf{0.440} & \textbf{0.387} & \textbf{94.3} \\
\bottomrule
\end{tabularx}
\captionof{table}{\textbf{\method also leads on standard metrics at multi-minute horizons.} Parcel-wise Pearson $r$ and whole-cortex global $r$ averaged over steps 1--20 or 1--40, and 20-step segment identification accuracy Cls@Top10 (\%, chance $0.05\%$, \textbf{best} and \underline{second best} per history length).}
\label{tab:standard}
\par
\endgroup
\end{appendixblock}

For each recording and lead step, \emph{Parcel $r$} correlates predicted and measured fMRI across anchor steps within each parcel before averaging over parcels, lead steps (1--20 or 1--40, where 1--40 equals $r$ in \cref{tab:main}), and recordings; \emph{Global $r$} computes one Pearson correlation jointly across anchor steps and parcels.
\emph{Cls@Top10} correlates each 20-step forecast against all $\sim 19{,}300$ 20-step segments from other recordings of the same individual, counting a hit when the ground-truth segment ranks in the top 10.
\method leads every metric at 120\,s and 180\,s, while at 7.5\,s MIRAGE leads on parcel $r$ and global $r$ and \method leads on Cls@Top10.

\paragraph{Correlation on \emph{State}.}
\emph{State} can also be scored with a correlation, which does not depend on the amplitude of the predictions.
At 180\,s, the Pearson correlation between the predicted and measured trial-to-trial differences $\hat d_t$ and $d_t$, pooled over parcels, steps, and evaluation windows, is $0.235$ for \method, $0.171$ for the linear forecaster, $0.118$ for NeuroWorld, $0.110$ for IPSID, and $0.088$ for BrainVista.

\begin{appendixblock}
\subsection{Transfer to the Human Connectome Project}
\label{sec:app-hcp}


\Cref{tab:hcp,tab:hcp-standard} evaluate \method and all baselines on Human Connectome Project (HCP) movie-fMRI data with 180\,s of history.

\medskip
\begingroup
\centering
\small
\begin{tabularx}{\linewidth}{>{\raggedright\arraybackslash}X*{3}{>{\centering\arraybackslash}X}}
\toprule
Model & \emph{Forecast}$\uparrow$ & \emph{State}$\uparrow$ & \emph{Updating}$\uparrow$ \\
\midrule
TRIBE v2~\cite{dascoli2026tribev2} & 2.6 [2.4, 2.9] & 0.0 & 2.5 [2.3, 2.7] \\
MIRAGE~\cite{gokce2026mirage} & 3.6 [3.2, 3.9] & 0.0 & 3.4 [3.0, 3.7] \\
Linear forecaster & \underline{6.2 [5.8, 6.5]} & \underline{5.1 [4.1, 6.0]} & 11.6 [11.1, 12.0] \\
IPSID~\cite{vahidi2024ipsid} & 4.7 [4.3, 5.0] & 5.0 [4.0, 6.1] & 12.8 [12.3, 13.4] \\
NeuroWorld~\cite{dong2026neuroworld} & 1.8 [1.5, 2.2] & 2.5 [1.3, 3.6] & \underline{13.1 [12.5, 13.8]} \\
BrainVista~\cite{yin2026brainvista} & 1.0 [0.9, 1.0] & 1.6 [1.1, 2.1] & 3.1 [3.0, 3.2] \\
\midrule
\rowcolor{rowhighlight}
\textbf{\method} & \textbf{6.6 [6.2, 7.0]} & \textbf{6.2 [4.7, 7.9]} & \textbf{15.7 [15.1, 16.3]} \\
\bottomrule
\end{tabularx}
\captionof{table}{\textbf{\method leads across all three \bench tasks on HCP.} Scores (\%) on held-out HCP movie clips (180\,s history) with 95\% confidence intervals from 2,000 bootstrap resamples over individuals (\textbf{best}, \underline{second best}).}
\label{tab:hcp}
\par
\endgroup

\medskip
\begingroup
\centering
\footnotesize
\setlength{\tabcolsep}{4.5pt}
\begin{tabularx}{\textwidth}{>{\raggedright\arraybackslash}Xccccc}
\toprule
 & \multicolumn{2}{c}{Parcel $r\uparrow$} & \multicolumn{2}{c}{Global $r\uparrow$} & Cls@Top10$\uparrow$ \\
\cmidrule(lr){2-3} \cmidrule(lr){4-5}
Model & 1--30 & 1--60 & 1--30 & 1--60 & (\%) \\
\midrule
TRIBE v2~\cite{dascoli2026tribev2} & 0.099 [0.093, 0.106] & 0.094 [0.088, 0.100] & 0.141 [0.134, 0.148] & 0.136 [0.130, 0.143] & 66.2 [63.9, 68.4] \\
MIRAGE~\cite{gokce2026mirage} & 0.125 [0.118, 0.131] & 0.121 [0.115, 0.127] & 0.168 [0.160, 0.176] & 0.163 [0.155, 0.171] & 76.9 [74.5, 79.2] \\
Linear forecaster & \underline{0.202 [0.197, 0.209]} & \underline{0.157 [0.152, 0.163]} & \underline{0.256 [0.250, 0.263]} & \underline{0.208 [0.202, 0.214]} & \underline{94.0 [93.2, 94.7]} \\
IPSID~\cite{vahidi2024ipsid} & 0.187 [0.182, 0.193] & 0.143 [0.138, 0.148] & 0.243 [0.237, 0.250] & 0.197 [0.191, 0.202] & 90.7 [89.7, 91.7] \\
NeuroWorld~\cite{dong2026neuroworld} & 0.164 [0.159, 0.170] & 0.109 [0.104, 0.113] & 0.219 [0.212, 0.227] & 0.156 [0.150, 0.163] & 86.5 [85.3, 87.8] \\
BrainVista~\cite{yin2026brainvista} & 0.099 [0.095, 0.104] & 0.055 [0.051, 0.058] & 0.137 [0.132, 0.143] & 0.094 [0.090, 0.098] & 65.2 [63.2, 67.4] \\
\midrule
\rowcolor{rowhighlight}
\textbf{\method} & \textbf{0.217 [0.211, 0.224]} & \textbf{0.163 [0.157, 0.169]} & \textbf{0.286 [0.279, 0.295]} & \textbf{0.229 [0.221, 0.236]} & \textbf{94.9 [94.1, 95.7]} \\
\bottomrule
\end{tabularx}
\captionof{table}{\textbf{\method also leads on standard metrics on HCP.} Parcel-wise Pearson $r$ and whole-cortex global $r$ averaged over steps 1--30 or 1--60, and 30-step segment identification accuracy Cls@Top10 (\%) on the forecasts of \cref{tab:hcp}, with 95\% confidence intervals from 2,000 bootstrap resamples over individuals (\textbf{best}, \underline{second best}).}
\label{tab:hcp-standard}
\par
\endgroup
\end{appendixblock}

\paragraph{Protocol and baseline adaptations.}
Of the 184 HCP individuals, models train on 148 individuals ($\approx 45$\,min of training movie data each) and are evaluated on their held-out clips, while \emph{State} is scored on a clip viewed twice by 27 of the 36 held-out individuals.
All step-based durations convert to HCP's 1\,s sampling rate (180-step history, 60-step rollout, 3\,s hemodynamic lag; parcel $r$ and global $r$ averaged over steps 1--30 or 1--60, and Cls@Top10 matching 30-step segments against $\sim 116$ candidates from other clips of the same individual), with penalties and checkpoints selected on held-out validation clips.
\method freezes its Algonauts-trained 5-step fast encoder $g_\phi$ (spanning 5\,s) and fits only the slow weights and linear predictor on HCP.
Given the shorter per-individual recordings, the linear forecaster, IPSID (32 states), NeuroWorld (35 history tokens covering 178\,s, 24-step rollouts, per-individual decoder heads; constrained to 3,533 contiguous 202-step windows across short 1--4.5\,min clips vs.\ 104,608 on Algonauts), and BrainVista (per-individual normalization; video and audio features only) each fit a single shared model across the 148 training individuals.
TRIBE~v2 runs unchanged; because MIRAGE's released heads do not cover HCP individuals (averaging them scores $0.6\%$ on \emph{Forecast}) and HCP lacks official transcripts, MIRAGE extracts features using Whisper word-timed transcripts~\cite{radford2023whisper}, linearly interpolates its 3,072-D trunk output from 1.49\,s to 1\,s, and fits a shared ridge head on training clips via leave-one-clip-out cross-validation.

\paragraph{Results.}
\method ranks first on every metric in \cref{tab:hcp,tab:hcp-standard}, while the linear forecaster ranks second on all metrics except \emph{Updating} (where NeuroWorld is second).
In paired bootstrap comparisons across individuals, \method outperforms every baseline significantly on all three \bench tasks and 60-step Pearson $r$, except against the linear forecaster on \emph{State}.

\begin{appendixblock}
\subsection{Probe calibration for unseen HCP individuals}
\label{sec:app-newviewer}

On the HCP individuals unseen during training, scored on their fourth recording, which is held out from fitting for every individual, \method is read out via a single shared linear probe and reaches $7.28\%$ on \emph{Forecast} and $16.70\%$ on \emph{Updating}.
\Cref{tab:newviewer} tests how much per-individual calibration adds by fitting a ridge residual correction to this shared probe on the initial minutes of each individual's first recording (penalty selected on training individuals).

\medskip
\begingroup
\centering
\small
\setlength{\tabcolsep}{6pt}
\begin{tabularx}{\linewidth}{>{\raggedright\arraybackslash}Xcccc}
\toprule
 & \multicolumn{2}{c}{Score (\%)} & \multicolumn{2}{c}{Gain over shared probe (points)} \\
\cmidrule(lr){2-3}\cmidrule(lr){4-5}
Calibration fMRI (min) & \emph{Forecast}$\uparrow$ & \emph{Updating}$\uparrow$ & \emph{Forecast} & \emph{Updating} \\
\midrule
\textbf{0 (shared probe)} & 7.28 & 16.70 & -- & -- \\
\midrule
1 & 7.29 & 16.71 & $+$0.01 [0.00, 0.02] & $+$0.01 [0.00, 0.03] \\
2 & 7.32 & 16.74 & $+$0.04 [0.01, 0.07] & $+$0.04 [$-$0.01, 0.09] \\
5 & 7.40 & 16.85 & $+$0.13 [0.07, 0.18] & $+$0.15 [0.04, 0.26] \\
15 & 7.67 & 17.13 & $+$0.39 [0.29, 0.49] & $+$0.43 [0.19, 0.73] \\
\bottomrule
\end{tabularx}
\captionof{table}{\textbf{Shared linear probe transfers zero-shot to unseen HCP individuals, requiring minimal calibration.} Scores (\%) on the held-out fourth recording of the unseen HCP individuals (35 for \emph{Forecast}, 34 for \emph{Updating}; 180\,s history) when adding a ridge correction fitted on the first 1--15\,min of each individual's first recording. Gains are paired differences computed before rounding with 95\% confidence intervals.}
\label{tab:newviewer}
\par
\endgroup
\end{appendixblock}

Fifteen minutes of personal calibration adds $+0.39$ points on \emph{Forecast} and $+0.43$ points on \emph{Updating}, showing that the shared linear probe already captures nearly all predictable signal on unseen individuals.

\begin{appendixblock}
\section{Ablation and long-horizon analyses}

This section expands \cref{sec:ablations} across forecast horizons and five-minute open-loop rollouts.

\subsection{Ablations across forecast horizons}
\label{sec:app-ablations}

\Cref{tab:app-ablations} breaks down the \emph{Forecast} ablations of \cref{tab:ablations} into near-term (0--7.5\,s) and far-term (30--60\,s) horizons.

\medskip
\begingroup
\centering
\small
\setlength{\tabcolsep}{7pt}
\begin{tabular}{@{}lccc@{}}
\toprule
 & \multicolumn{3}{c}{\emph{Forecast} $\Delta$ (points), horizon} \\
\cmidrule(l){2-4}
Variant & 0--60\,s & 0--7.5\,s & 30--60\,s \\
\midrule
\multicolumn{4}{@{}l}{\textit{(a) Architecture}} \\
w/o slow state $u_t$ & $-4.6$ & $-9.5$ & $-1.7$ \\
Slow weights: learned $\to$ mean & $-2.8$ & $-6.4$ & $-1.3$ \\
Transformer encoder $g_\phi$ & $-1.2$ & $-3.8$ & $-0.5$ \\
Transformer predictor $p_\theta$ & $-1.2$ & $-0.6$ & $-1.3$ \\
Transformer $g_\phi$ and $p_\theta$ & $-2.7$ & $-5.3$ & $-1.7$ \\
\addlinespace
\multicolumn{4}{@{}l}{\textit{(b) Training}} \\
w/o pretraining (random init.) & $-2.9$ & $-7.3$ & $-1.1$ \\
Frozen pretrained encoder & $-0.7$ & 0.0 & $-1.1$ \\
w/o EMA and stop-gradient & $-0.8$ & $-0.2$ & $-1.2$ \\
\addlinespace
\multicolumn{4}{@{}l}{\textit{(c) Stimulus modalities}} \\
w/o video features & $-2.5$ & $-1.1$ & $-2.9$ \\
w/o text features & $-1.1$ & $-0.3$ & $-1.4$ \\
w/o audio features & $-0.6$ & $-0.4$ & $-0.6$ \\
\addlinespace
\multicolumn{4}{@{}l}{\textit{(d) Hyperparameters}} \\
Fast-state window: 7.5 $\to$ 4.5\,s & $-0.1$ & $-0.2$ & 0.0 \\
Stimulus window: 7.5 $\to$ 1.5\,s & $-0.5$ & $-0.6$ & $-0.4$ \\
Stimulus window: 7.5 $\to$ 15\,s & $-0.1$ & 0.0 & $-0.1$ \\
State size: 2,048 $\to$ 1,024 & $-0.2$ & $-0.5$ & $-0.2$ \\
State size: 2,048 $\to$ 4,096 & $+0.1$ & $+0.5$ & $-0.1$ \\
Rollout $K$: 40 $\to$ 80 & $-0.1$ & 0.0 & $-0.1$ \\
\bottomrule
\end{tabular}
\captionof{table}{\textbf{Encoder design governs near-term accuracy while latent rollout training sustains long-horizon forecasts.} Change in \emph{Forecast} (percentage points relative to full \method) across the full minute (0--60\,s), near future (0--7.5\,s), and far future (30--60\,s).}
\label{tab:app-ablations}
\par
\endgroup
\end{appendixblock}

Dropping slow state $u_t$, encoder pretraining, or the linear encoder $g_\phi$ primarily degrades near-term predictions (0--7.5\,s), whereas freezing the pretrained encoder, removing EMA/stop-gradient targets, dropping video or text features, or replacing $p_\theta$ with a Transformer primarily hurts 30--60\,s accuracy.

\begin{appendixblock}
\subsection{Five-minute rollouts}
\label{sec:app-long}

\Cref{fig:long} extends open-loop \emph{Forecast} rollouts from one to five minutes without retraining in the same recording.

\medskip
\begingroup
\centering
\includegraphics[width=\linewidth]{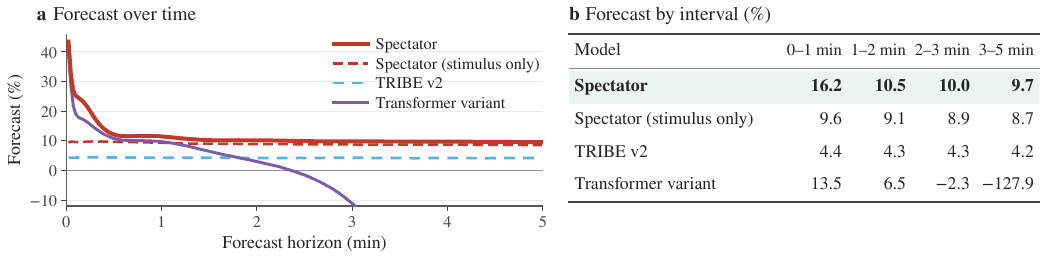}
\captionof{figure}{\textbf{\method stays stable for five minutes, whereas its Transformer variant diverges.} (a) Per-step \emph{Forecast} accuracy over 5\,min. (b) Mean scores (\%) within each horizon interval. Stimulus-only \method and TRIBE v2~\cite{dascoli2026tribev2} receive no fMRI history; the Transformer variant replaces both $g_\phi$ and $p_\theta$ (\cref{tab:ablations}), and its trajectory continues below the plotted range.}
\label{fig:long}
\par
\endgroup
\end{appendixblock}

\method remains above its stimulus-only baseline across all four intervals (0--1, 1--2, 2--3, and 3--5\,min), leading by $+6.6$, $+1.4$, $+1.1$, and $+1.0$ points ($95\%$ CI $[0.7, 1.3]$ at 3--5\,min).
The latent-state component accounts for $38\%$ of forecast variance in minute 0--1 and under $1\%$ past minute 3, concentrating most state-driven gain within the first minute.
By contrast, the Transformer $g_\phi$ and $p_\theta$ variant drops below the stimulus-only baseline after $67$\,s and diverges below zero after $143$\,s.

\begin{appendixblock}
\section{Latent state and stimulus contributions}
\label{sec:app-analysis}

The following analyses extend \cref{sec:analysis}.

\paragraph{Exact forecast decomposition.}
Partition the predictor weights as $W = [W_s,\ W_a,\ b]$, where $W_s$ acts on the latent state, $W_a$ acts on the stimulus features, and $b$ is the bias.
With $M = I + W_s$, unrolling the residual update in \cref{eq:predictor} gives
\begin{equation}
\hat s_{t+k}
= \underbrace{M^k s_t}_{\text{initial state}}
+ \underbrace{\sum_{j=1}^{k} M^{k-j}(W_a \bar a_{t+j} + b)}_{\text{stimulus (incl.\ bias)}}.
\label{eq:state-stimulus-decomposition}
\end{equation}
We group the bias $b$ with the stimulus term by convention.
Because the evaluation probe (\cref{sec:probe}) is also linear, mapping $\hat s_{t+k}$ to observation space preserves this exact additive split in $\hat x_{t+k}$, allowing either component to be evaluated, omitted, or swapped across time or individuals without rerunning the rollout.

\subsection{Swapping state and stimulus across time and individuals}
\label{sec:app-swap}


\Cref{tab:swap} reports \emph{Forecast} accuracy when the initial-state and stimulus terms each come from the anchor step, are omitted, or are replaced from midway through the same recording.

\medskip
\begingroup
\centering
\small
\setlength{\tabcolsep}{8pt}
\begin{tabular}{@{}lccc@{}}
\toprule
 & \multicolumn{3}{c}{Stimulus component} \\
\cmidrule(l){2-4}
Latent-state component & Original & Omitted & Another time \\
\midrule
Original & \textbf{16.0} & 7.5 & $-$1.4 \\
Omitted & 9.4 & 0.0 & $-$8.9 \\
Another time & 2.4 & $-$7.0 & $-$16.9 \\
\bottomrule
\end{tabular}
\captionof{table}{\textbf{Mismatched state or stimulus inputs actively degrade forecasts below omitting them.} \emph{Forecast} accuracy (\%) across all $3 \times 3$ combinations of keeping, omitting, or swapping \method's inferred latent state and stimulus with inputs from midway through the same recording.}
\label{tab:swap}
\par
\endgroup
\end{appendixblock}

Substituting either component from another time point consistently scores below omitting that component altogether.
The same holds when substituting another individual's state at the identical movie moment ($6.7\%$ vs.\ $9.4\%$ without state) or the same individual's state from the alternate viewing of a repeated movie ($4.5\%$ vs.\ $13.7\%$ with the original state and $7.4\%$ without state on repeated movies), whereas averaging the states of the three other individuals at that moment reaches $9.7\%$.

\paragraph{Input conditions of \cref{tab:inputs}.}
All rows of \cref{tab:inputs} use the same fitted \method and the same probes, and nothing is refitted.
\emph{Without stimulus} drops the stimulus term from every predicted step.
\emph{Without fMRI} replaces every fMRI step that the encoder reads with zeros, both in the history before $t_0$ and at the steps revealed in \emph{Updating}.
\emph{Another individual's fMRI} replaces the same steps with the fMRI of a different individual at the same moments of the same movie, and the forecasts are scored against the original individual's fMRI through the original individual's probe.
The \emph{Forecast} scores of the first two conditions equal the corresponding cells of \cref{tab:swap} ($7.5\%$ and $9.4\%$).

\subsection{Cortical gain maps and input attribution}
\label{sec:app-spatial}
\label{sec:app-cortex}

\paragraph{Gain maps on the cortex.}
In each parcel, stimulus gain is full \emph{Forecast} accuracy minus state-only accuracy, and latent-state gain is full accuracy minus stimulus-only accuracy across the anchor steps of \cref{tab:swap}.
For display, \cref{fig:cortex} averages the 1,000 parcels into the 100-region Schaefer parcellation~\cite{schaefer2018parcellation} and labels the top three externally visible regions with distinct NeuroQuery terms~\cite{dockes2020neuroquery} (selected from 20 candidate functional terms: auditory, speech perception, story, visual perception, visual motion, faces, reading, language comprehension, spatial attention, working memory, cognitive control, memory retrieval, theory of mind, self referential, motor, touch, pain, reward, emotion, and decision making).
Inter-subject correlation in \cref{fig:cortex}c is the parcel-wise correlation between each individual's fMRI and the mean of the other individuals on the same recording, averaged across individuals and recordings~\cite{hasson2004intersubject}.

\begin{appendixblock}
\paragraph{Parcels the forecast relies on.}
\label{sec:app-relies}
To identify which historical regions drive forecasts, we compute exact GradientSHAP attributions~\cite{lundberg2017unified} from the 180\,s fMRI history onto predicted parcels.
Because \method is linear in past fMRI, the attribution of forecast parcel $j$ to input parcel $i$ equals the exact input gradient multiplied by the fMRI difference relative to a reference anchor step, summed over the 180\,s history; overall input importance sums absolute attributions across forecast parcels and averages over 300 anchor pairs at lead times of 1.5, 15, and 60\,s.
\Cref{fig:relies} shows the top 100 input parcels, led by default-mode (38), frontoparietal control (20), and visual (19) networks~\cite{yeo2011organization}, including the angular gyrus and precuneus.

\medskip
\begingroup
\centering
\includegraphics[width=\linewidth]{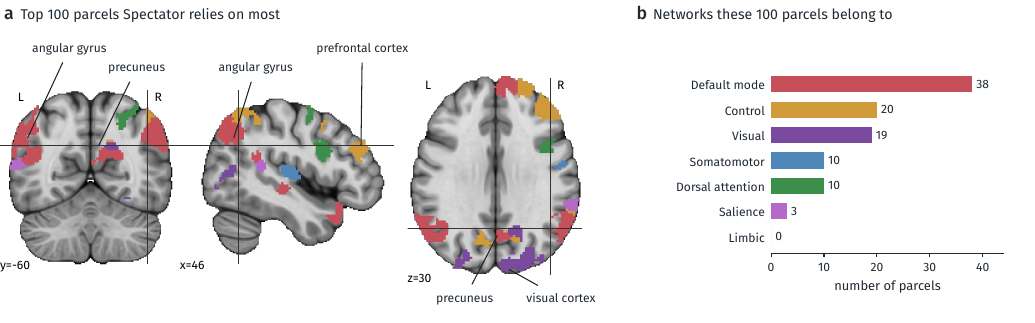}
\captionof{figure}{\textbf{Forecasts rely primarily on transmodal default-mode and control networks.} (a) Top 100 input parcels ranked by GradientSHAP attribution of past fMRI observations onto \method's forecasts, colored by functional network~\cite{yeo2011organization}. (b) Network breakdown, led by default-mode (38) and frontoparietal control (20) regions.}
\label{fig:relies}
\par
\endgroup
\end{appendixblock}

\begin{appendixblock}
\subsection{Learned multi-timescale dynamics}
\label{sec:app-dynamics}

\Cref{fig:inside} characterizes the temporal balance between state and stimulus, the model's effective impulse response to stimulus perturbations, and the decay spectrum of its latent transition matrix $M$.

\medskip
\begingroup
\centering
\includegraphics[width=\linewidth]{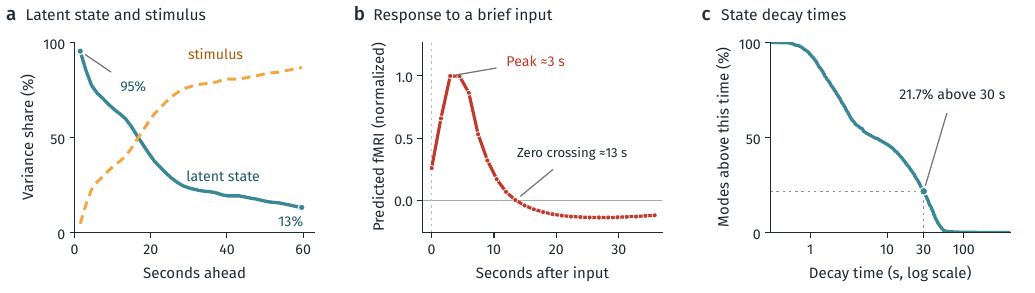}
\captionof{figure}{\textbf{\method learns stable, multi-timescale latent dynamics.} (a) Relative variance shares of the initial-state and stimulus components across forecast horizons. (b) Normalized fMRI impulse response to a single-step stimulus perturbation across responsive parcels, peaking near 3\,s. (c) Survival distribution of eigenmode $1/e$ decay times across all 2,048 latent modes, with 21.7\% persisting beyond 30\,s.}
\label{fig:inside}
\par
\endgroup
\end{appendixblock}

\paragraph{Latent state and stimulus.}
At each lead step, we compute the variance of the state and stimulus components across anchor steps, sum across parcels, and normalize by their combined variance (excluding covariance).
The initial-state share drops from $95\%$ at 1.5\,s to $13\%$ at 60\,s (\cref{fig:inside}a, matching \cref{fig:time}a) as unforced state modes decay while incoming stimuli continually drive the rollout.

\paragraph{Response to a brief input.}
We propagate a single-step stimulus perturbation along the principal input direction (maximizing response norm under the summed five-lag input-to-parcel maps for one individual) through the input map, transition $M$, and linear probe.
Averaging the 50 parcels with the largest peak absolute responses after peak-sign alignment and peak normalization yields a response that peaks near 3\,s, crosses zero near 13\,s, and shows a small negative undershoot (\cref{fig:inside}b), reflecting both the 5-step stimulus window and recurrent propagation.

\paragraph{State decay times.}
Each eigenvalue $\lambda_i$ of $M$ defines a latent eigenmode with $1/e$ decay time $\tau_i = -\Delta t / \log|\lambda_i|$ ($\Delta t = 1.49$\,s).
All 2,048 eigenvalues satisfy $|\lambda_i| < 1$ (maximum $0.9951$), ensuring asymptotic stability of unforced rollouts, while 444 modes ($21.7\%$) persist beyond 30\,s (\cref{fig:inside}c) to sustain multi-minute context alongside stimulus-driven updates.

\begin{appendixblock}
\subsection{Dimensionality of the latent state}
\label{sec:app-dim}

\Cref{fig:dim} shows how many dimensions of the 2,048-dimensional latent state the linear probe needs.

\medskip
\begingroup
\centering
\includegraphics[width=\linewidth]{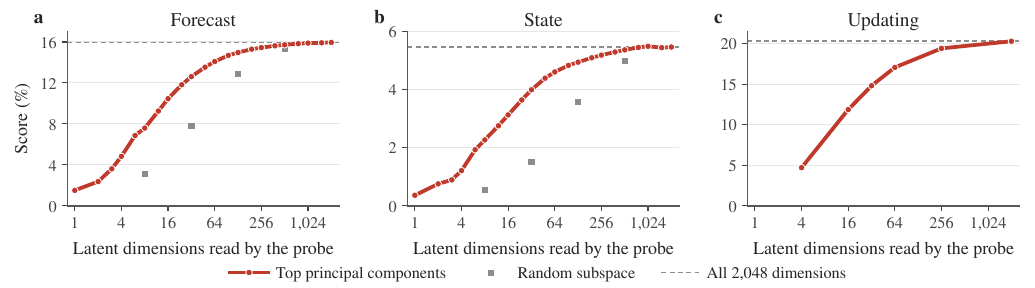}
\captionof{figure}{\textbf{About 100 of the 2,048 latent dimensions recover 90\% of \emph{Forecast} and \emph{State}.} Scores on (a) \emph{Forecast}, (b) \emph{State}, and (c) \emph{Updating} when each individual's probe reads only the top principal components of the predicted latent states, or a random subspace of the same size. \method itself is unchanged. Dashed lines mark the probe on all 2,048 dimensions (\cref{tab:main}).}
\label{fig:dim}
\par
\endgroup
\end{appendixblock}

\method and its rollouts are kept unchanged.
For each individual, we refit the linear probe on the top $k$ principal components of the latent states that \method predicts on the training recordings, with the ridge penalty selected as for the full probe.
In \emph{Updating}, the same truncated probe also imputes the unobserved steps.
Twelve components recover half of the full \emph{Forecast} and \emph{State} scores, and 48 recover $80\%$.
The scores reach $90\%$ with 96 components on \emph{Forecast} and with 128 on \emph{State}, and \emph{Updating} reaches $84\%$ with 64 components and $96\%$ with 256.
A random subspace of the same size recovers less: 32 random dimensions give $1.5\%$ on \emph{State}, compared with $4.0\%$ for the top 32 components.
The information that the probe reads about an individual's brain state is therefore concentrated in about one hundred directions of the latent state.

\begin{appendixblock}
\section{Data efficiency and trajectory distillation}
\label{sec:app-extra}

This section expands the data-efficiency analysis in \cref{sec:main,sec:discussion} and evaluates downstream student models trained on \method's simulated trajectories.
All models are scored on the held-out Algonauts split of \cref{tab:main} with 180\,s of fMRI history and 95\% confidence intervals from 2,000 paired bootstrap resamples (\cref{sec:app-significance}).

\subsection{Scaling with training-data size}
\label{sec:app-datasize}

\Cref{fig:data,tab:datasize} report performance across nested subsets of the 118 \emph{Friends} training episodes (238 fMRI recordings, $46.6$\,h per individual).

\medskip
\begingroup
\centering
\includegraphics[width=\linewidth]{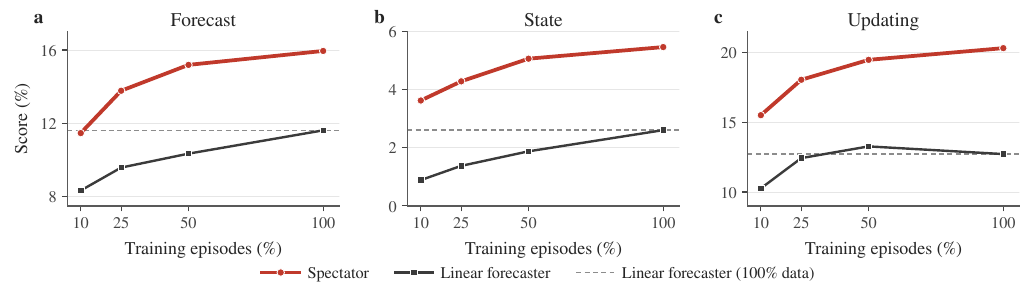}
\captionof{figure}{\textbf{\method scales effectively from limited per-individual fMRI.} Performance on (a) \emph{Forecast}, (b) \emph{State}, and (c) \emph{Updating} on held-out fMRI (180\,s history) across matched nested subsets of the 118 training episodes per individual (\method retrained end-to-end at each budget). Dashed horizontal lines mark the linear forecaster trained on 100\% of the episodes.}
\label{fig:data}
\par
\endgroup

\medskip
\begingroup
\centering
\small
\setlength{\tabcolsep}{3.6pt}
\begin{tabularx}{\linewidth}{>{\centering\arraybackslash}X>{\centering\arraybackslash}Xccccccccc}
\toprule
 & & \multicolumn{3}{c}{\textbf{\method}} & \multicolumn{6}{c}{Linear forecaster} \\
\cmidrule(lr){6-11}
\multicolumn{2}{c}{Training budget} & \multicolumn{3}{c}{(retrained end-to-end)} & \multicolumn{3}{c}{Recorded fMRI targets} & \multicolumn{3}{c}{\method forecast targets} \\
\cmidrule(lr){1-2}\cmidrule(lr){3-5}\cmidrule(lr){6-8}\cmidrule(lr){9-11}
Data (\%) & Episodes (h) & \emph{Forecast}$\uparrow$ & \emph{State}$\uparrow$ & \emph{Updating}$\uparrow$ & \emph{Forecast}$\uparrow$ & \emph{State}$\uparrow$ & \emph{Updating}$\uparrow$ & \emph{Forecast}$\uparrow$ & \emph{State}$\uparrow$ & \emph{Updating}$\uparrow$ \\
\midrule
10\% & 12 (4.7) & \textbf{11.5} & \textbf{3.6} & \textbf{15.5} & 8.3 & 0.9 & 10.3 & 9.1 & 1.6 & 12.0 \\
25\% & 30 (11.7) & \textbf{13.8} & \textbf{4.3} & \textbf{18.1} & 9.6 & 1.4 & 12.4 & 11.3 & 3.0 & 13.1 \\
50\% & 59 (23.3) & \textbf{15.2} & \textbf{5.1} & \textbf{19.5} & 10.4 & 1.9 & 13.3 & 12.5 & 4.0 & 15.3 \\
100\% & 118 (46.6) & \textbf{16.0} & \textbf{5.5} & \textbf{20.3} & 11.6 & 2.6 & 12.7 & 13.6 & 4.7 & 16.5 \\
\bottomrule
\end{tabularx}
\captionof{table}{\textbf{\method leads across all metrics at every training-data budget.} Scores (\%) on held-out fMRI (180\,s history) for matched nested subsets of the 118 training episodes ($46.6$\,h per individual). \method is retrained end-to-end on each subset; the linear forecaster fits either to recorded fMRI or to forecasts generated by the \method teacher trained on the same subset. Bold marks the best score at each data budget.}
\label{tab:datasize}
\par
\endgroup
\end{appendixblock}

With 10\% of the episodes ($\approx 4.7$\,h per individual), \method already surpasses the full-data linear forecaster on \emph{State} ($+1.0$ [$0.9$, $1.1$]) and \emph{Updating} ($15.5\%$ vs.\ $12.7\%$) while matching its \emph{Forecast} score within $0.15$ points ($-0.15$ [$-0.27$, $-0.02$]).
With 25\% ($\approx 11.7$\,h), \method beats the full-data linear forecaster on all three tasks ($+2.2$ points [$2.1$, $2.3$] on \emph{Forecast}).
At matched budgets (10\%, 25\%, 50\%, 100\%), \method leads the recorded-fMRI linear forecaster by $3.1$, $4.2$, $4.8$, and $4.3$ points on \emph{Forecast}, $2.7\text{--}3.2$ points on \emph{State}, and $5.2\text{--}7.6$ points on \emph{Updating}, continuing to improve from 50\% to 100\% ($+0.8$ points on \emph{Forecast} and \emph{Updating}) as its train--test \emph{Forecast} gap narrows from $7.2$ points at 10\% to $1.8$ points at 100\% (vs.\ $5.0$ points for the full-data linear forecaster).

\begin{appendixblock}
\subsection{Learning from \method's trajectories}
\label{sec:app-generated}

Beyond direct forecasting, a world model can supervise downstream learners by supplying denoised rollout targets on observed sequences and counterfactual trajectories under novel state--stimulus pairings.
\Cref{tab:app-generated} evaluates three student families trained on \method's trajectories and scored strictly on real held-out fMRI.

\medskip
\begingroup
\centering
\footnotesize
\setlength{\tabcolsep}{3.5pt}
\renewcommand{\arraystretch}{1.0}
\begin{tabularx}{\linewidth}{l>{\raggedright\arraybackslash}Xcccccc}
\toprule
 & & \multicolumn{3}{c}{Score (\%)} & \multicolumn{3}{c}{Gain over real targets [95\% CI]} \\
\cmidrule(lr){3-5}\cmidrule(lr){6-8}
Student model & Training targets & \emph{Forecast}$\uparrow$ & \emph{State}$\uparrow$ & \emph{Updating}$\uparrow$ & \emph{Forecast} & \emph{State} & \emph{Updating} \\
\midrule
\textbf{Linear forecaster} & Recorded fMRI & 11.6 & 2.6 & 12.7 & -- & -- & -- \\
 & Equal mix (recorded + forecasts) & 12.8 & 3.8 & 15.5 & $+$1.2 [1.2, 1.2] & $+$1.2 [1.2, 1.3] & $+$2.8 [2.8, 2.9] \\
\rowcolor{rowhighlight}
 & \method forecasts & \textbf{13.6} & \textbf{4.7} & \textbf{16.5} & $+$2.0 [1.9, 2.0] & $+$2.1 [2.0, 2.2] & $+$3.8 [3.7, 3.8] \\
\addlinespace[3pt]
\textbf{\method (Transformer $p_\theta$)} & Encoded fMRI & 14.5 & 4.4 & 19.2 & -- & -- & -- \\
 & + \method forecasts & 15.4 & 5.1 & 19.5 & $+$0.9 [0.9, 0.9] & $+$0.8 [0.7, 0.8] & $+$0.3 [0.3, 0.4] \\
\rowcolor{rowhighlight}
 & + Counterfactual pairs & \textbf{15.5} & \textbf{5.3} & \textbf{19.8} & $+$1.0 [1.0, 1.1] & $+$0.9 [0.9, 1.0] & $+$0.5 [0.5, 0.6] \\
\addlinespace[3pt]
\textbf{NeuroWorld}~\cite{dong2026neuroworld} & Recorded fMRI & 8.3 & 0.4 & 13.4 & -- & -- & -- \\
\rowcolor{rowhighlight}
 & + \method forecasts & \textbf{9.4} & \textbf{1.5} & \textbf{13.9} & $+$1.2 [1.1, 1.3] & $+$1.0 [0.9, 1.2] & $+$0.6 [0.5, 0.6] \\
\midrule
\emph{\method (teacher)} & & 16.0 & 5.5 & 20.3 & & & \\
\bottomrule
\end{tabularx}
\captionof{table}{\textbf{Supervising downstream models with \method's trajectories yields significant gains across all students.} Scores (\%) on real held-out fMRI (180\,s history) and paired differences (percentage points, computed before rounding) relative to each student's real-target baseline with 95\% bootstrap confidence intervals. Equal mix averages recorded fMRI and \method forecast targets; ``+'' augments the preceding target set. Shading and bold mark the best setting per student.}
\label{tab:app-generated}
\par
\endgroup
\end{appendixblock}

\paragraph{Training targets.}
All students read real fMRI history and stimulus features while \method supplies rollout targets.
The linear forecaster fits to \method's forecast from the same anchor step or its equal average with recorded fMRI.
For the 6-block Transformer $p_\theta$ on \method's frozen encoder (16-step rollouts, scoring $14.5\%$ on \emph{Forecast} vs.\ $14.8\%$ at 40 steps in \cref{tab:ablations}), next-step and rollout losses add equal-weight matching losses against \method's latent rollouts, plus counterfactual rollouts pairing one window's latent state with another's stimulus.
For NeuroWorld (trained on a single GPU under the same hyperparameters as \cref{tab:main}), each training window is duplicated with its last 16 steps replaced by \method's forecast.

\paragraph{Results.}
All three students improve significantly under \method supervision while \method remains strongest (\cref{tab:app-generated}), showing that its rollouts suppress fMRI measurement noise while preserving personal state signal.
Counterfactual pairs improve the Transformer student by $+0.13$ points [$0.12$, $0.15$] on \emph{Forecast} beyond in-distribution rollouts and narrow its train--test \emph{Forecast} gap from $2.9$ to $1.7$ points (matching \method's $1.8$, whereas NeuroWorld's rises from $3.0$ to $4.2$).
At 10\%, 25\%, 50\%, and 100\% data, distilling a same-subset \method teacher lifts the linear forecaster by $+0.7$, $+1.7$, $+2.2$, and $+2.0$ points on \emph{Forecast} (\cref{tab:datasize}); at 50\% data it beats the 100\% real-target forecaster on all three tasks ($+0.9$ [$0.9$, $1.0$], $+1.4$ [$1.3$, $1.5$], and $15.3\%$ vs.\ $12.7\%$).
With 7.5\,s history, distillation gives no gain ($-0.3$ alone; $0.0$ mixed), confirming that multi-minute state drives the improvement.

\begin{appendixblock}
\section{Additional related work}
\label{sec:app-related}

\paragraph{Medical world models.}
Medical imaging world models learn diagnostic representations~\cite{yue2025chexworld,li2026gazeworld}, guide ultrasound~\cite{yue2025echoworld}, simulate tumor progression~\cite{yang2025mewm}, synthesize contrast MRI~\cite{kong2026cekworld}, and generate CT reports~\cite{tian2026sliceworld}, evaluated via task accuracy, guidance error, Turing tests, image fidelity, or report metrics.
For biosignals, \citet{bao2026biosignals} further call for explicit intervention conditioning and causal verification; complementing clinical tasks, we model and evaluate ongoing brain state under naturalistic stimulation.

\paragraph{Latent prediction and world-model evaluation.}
Joint-embedding world models plan in latent space~\cite{zhou2025dinowm,assran2025vjepa2,sobal2025pldm,maes2026lewm}. High visual quality need not imply effective planning~\cite{shang2026worldarena,zhang2026worldinworld}, physical consistency~\cite{motamed2025physicsiq}, or accurate tracking of out-of-view state~\cite{lu2026wrbench}.
Unlike synthetic domains with ground-truth states~\cite{li2023othello,vafa2024evaluating}, brain states are unobserved, so \bench tests state inference through forecasting, repeated stimuli, and mid-rollout state updating.
\end{appendixblock}

\end{document}